\documentclass[journal,12pt,onecolumn,draftclsnofoot,]{IEEEtran}
\ifCLASSINFOpdf
\else
\fi
\usepackage{latexsym,amsfonts,amsmath,epsfig,amssymb}
\usepackage{graphicx,times}
\usepackage{mathrsfs}
\usepackage{caption}
\usepackage{subfigure}
\usepackage{graphicx}
\usepackage{epstopdf}
\usepackage{multirow}
\usepackage{booktabs}
\usepackage{color}
\usepackage{url}

\usepackage{algorithm,algpseudocode}

\usepackage[numbers,sort&compress]{natbib}

\graphicspath{{Figuressparse/}}                  

\renewcommand{\theequation}{\arabic{equation}}
\renewcommand{\thetable}{\arabic{table}}
\renewcommand{\thefigure}{\arabic{figure}}

\newtheorem{theorem}{Theorem}[section]

\newtheorem{Prop}{Proposition}[section]

\newtheorem{lemma}{Lemma}[section]

\newtheorem{remark}{Remark}[section]
\newtheorem{definition}{Definition}[section]

\newcounter{nextauthor}
\def\mathrm{\mbox}

\numberwithin{remark}{section}

\renewcommand{\baselinestretch}{1}  

\begin{document}
%
\title{Sparse Nonnegative Tensor Factorization and Completion with  Noisy Observations}
%
%
%

\author{Xiongjun~Zhang
        and~Michael K. Ng
\thanks{The work of X. Zhang was  partially supported by the
	National Natural Science Foundation of China under Grant Nos. 11801206 and 12171189.
The work of M. K. Ng was  partially supported by the HKRGC GRF
12300218, 12300519, 17201020
and 17300021.}
\thanks{X. Zhang is with the School of Mathematics and Statistics and Hubei Key Laboratory of Mathematical Sciences,
	Central China Normal University, Wuhan 430079, China (e-mail: xjzhang@mail.ccnu.edu.cn).}
\thanks{M. K. Ng is wth the Department of Mathematics, The University of
	Hong Kong, Pokfulam, Hong Kong (e-mail: mng@maths.hku.hk).}}

\maketitle

\begin{abstract}
In this paper, we study the sparse nonnegative tensor factorization and completion problem from partial
and noisy observations for third-order tensors. Because of sparsity and nonnegativity,
the underlying tensor is decomposed into the tensor-tensor product of one
sparse nonnegative tensor and one nonnegative tensor.
We propose to minimize the sum of the maximum likelihood
estimation for the observations with nonnegativity constraints and the tensor
$\ell_0$ norm for the sparse factor. We show that the error
bounds of the estimator of the proposed model can be established under general noise observations.
The detailed error bounds under specific noise distributions including additive Gaussian noise,
additive Laplace noise, and Poisson observations can be derived.
Moreover, the minimax lower bounds are shown to be matched with the established upper bounds
up to a logarithmic factor of the sizes of the underlying tensor. These theoretical results for tensors
are better than those obtained for matrices, and this illustrates the advantage of the use of
nonnegative sparse tensor models for completion and denoising.
Numerical experiments are provided to validate the superiority
of the proposed tensor-based method compared with the matrix-based approach.
\end{abstract}


%
\IEEEpeerreviewmaketitle

\section{Introduction}

\IEEEPARstart{W}{ith} the rapid development of computer technique,  multi-dimensional data,
which are also known as tensors \cite{kolda2009tensor}, have received  much attention
in various application fields, such as data mining \cite{kolda2008scalable, morup2011applications},
signal and  image processing \cite{cichocki2007nonnegative, cichocki2015tensor,  sidiropoulos2017tensor, zhang2019nonconvex},
and  neuroscience \cite{mori2006principles}. Many underlying tensor data are
nonnegative due to their physical meaning such as the pixels of images.
An efficient approach to exploit the intrinsic structure of a nonnegative tensor is tensor factorization,
which can  explore its hidden information.
Moreover, the underlying tensor data may also suffer from missing entries and noisy corruptions
during its acquiring process.
In this paper, we focus on the sparse nonnegative tensor factorization (NTF) and completion
problem from partial and noisy observations, where the observed entries are corrupted
by general noise distributions
such as additive Gaussian noise, additive Laplace noise, and Poisson observations.

Tensors arise in a variety of real-world applications that
can represent the multi-dimensional correlation of tensor data,
e.g., the spatial and spectral dimensions for hyperspectral images
and the spatial and time dimensions for video data.
In particular, for  second-order tensors,
NTF reduces to nonnegative matrix factorization (NMF),
which can extract meaningful features
and has a wide variety of practical applications in scientific and engineering areas,
see \cite{ding2008convex, lee1999learning, gillis2020nonnegative, pan2019generalized,  pan2018orthogonal}
and references therein.
It has been demonstrated that NMF is
able to learn localized features with obvious interpretations \cite{lee1999learning}.
Moreover,  Gillis \cite{gillis2012sparse} proposed a sparse NMF model with a sparse factor,
which provably led to optimal and sparse solutions under a separability assumption.
Recently, Soni et al. \cite{soni2016noisy} proposed a general class of
matrix completion tasks with noisy observations, which could reduce to
sparse NMF when the underlying factor matrices are nonnegative and all
entries of the noisy matrix are observed.
They showed that the error
bounds of estimators of sparse NMF are lower than those of NMF \cite{soni2016noisy}.
 More theory and applications about the advantages of sparse NMF over NMF can be referred to
 \cite{gao2005improving, zhi2010graph, gillis2010using, kim2007sparse,  soltani2017tomographic, sambasivan2018minimax}.


%

Due to  exploiting  the intrinsic structure of the underlying tensor data,
which contains correlation in different modes,
NTF has also been widely applied in a variety of fields,
see, e.g., \cite{ chi2012tensors, hong2020generalized,  pan2021orthogonal, veganzones2015nonnegative}.
There are some popular NTF approaches,
such as nonnegative Tucker decomposition \cite{li2016mr},
nonnegative CANDECOMP/PARAFAC (CP) decomposition \cite{veganzones2015nonnegative},
nonnegative tensor train decomposition \cite{lee2016nonnegative},
which are  derived by different
applications, see also \cite{kolda2009tensor, vervliet2019exploiting}.
For example, Xu \cite{xu2015alternating} proposed an  alternating proximal
gradient method for sparse nonnegative Tucker decomposition,
while it is only efficient for additive Gaussian noise.
 Other applications about sparse NTF can be found in \cite{qi2018muti, morup2008algorithms, kim2013sparse}.

Another kind of NTF is based on the recently proposed
tensor-tensor product \cite{Kilmer2011Factorization},
whose algebra operators have been proposed and studied
for third-order tensors \cite{Kilmer2011Factorization, Kilmer2013Third}
and then generalized to higher-order tensors \cite{martin2013order} and transformed tensor-tensor product \cite{song2020robust}.
Besides, Kilmer et al. \cite{Kilmer2011Factorization} established the framework of
tensor singular value decomposition (SVD) for third-order tensors.
This kind of tensor-tensor product and tensor SVD
has been applied in a great number of  areas such as facial recognition \cite{hao2013facial},
tensor completion \cite{ng2020patch, corrected2019zhang, zhang2014novel,Zhang2017Exact, zhang2021low},
and image processing \cite{semerci2014tensor, zheng2020mixed}.
Recently, this kind of sparse NTF models
has been proposed and studied
on dictionary learning problems, e.g.,
tomographic image reconstruction  \cite{soltani2016tensor},
image compression and image deblurring \cite{newman2019non}.
The sparse factor of NTF with tensor-tensor product is due to the sparse
representation of patched-dictionary elements for tensor dictionary learning \cite{soltani2016tensor}.
One needs to learn a nonnegative tensor patch dictionary from training data,
which is to solve a sparse NTF problem with tensor-tensor product. It was demonstrated that
the tensor-based dictionary
learning algorithm exhibits better performance
than the matrix-based method in terms of approximation accuracy.
However, there is no theoretical result about the error bounds of sparse NTF models.
Both different noise settings and missing values
are not studied in the literature.

In this paper, we propose a sparse NTF and completion model with tensor-tensor product
from partial and noisy observations for third-order tensors, where the observations
are corrupted by a general class of noise models.
The proposed model consists of a data-fitting term for the observations
and the tensor $\ell_0$ norm for the sparse factor,
where the two tensor factors operated by tensor-tensor product are nonnegative
and the data-fitting term is derived by maximum likelihood estimation.
Theoretically, we show that the error
bounds of the estimator of the proposed model can be established under general noise observations.
The detailed error bounds under specific noise distributions including additive Gaussian noise,
additive Laplace noise, and Poisson observations can be derived.
Moreover, the minimax lower bounds are shown to be matched with the established upper bounds
up to a logarithmic factor of the sizes of the underlying tensor. These theoretical results for tensors
are better than those obtained for matrices in \cite{soni2016noisy}, and this illustrates the advantage of the use of
nonnegative sparse tensor models for completion and denoising.
 In Table \ref{DiffMedC}, we summarize existing sparse NMF and  NTF results for an $n_1\times n_2\times n_3$ tensor.

\begin{table}[!t]
	\scriptsize
	\begin{center}
		\setlength{\abovecaptionskip}{-1pt}
		\setlength{\belowcaptionskip}{-1pt}
		\caption{\small  Comparisons of different methods for sparse nonnegative tensor factorization and completion.}\label{DiffMedC}
		\medskip
		\begin{tabular}{| c | c|c  c c   |} \hline
			     {Methods}  & Rank   & {Noise type}  &{Missing values} & {Error bounds}       \\
			 \hline \hline
			 Matrix based method \cite{soni2016noisy} & matrix $r$              & general noise  & Yes   & $O(\frac{rn_1n_3+\|\mathcal{B}\|_0}{m}\log(\max\{n_1n_3,n_2\}))$  \\
		     Sparse nonnegative CPD \cite{jain2017noisy}         & CP rank $r$      & general noise  & Yes   & $O(\frac{(n_1+n_2)r+\|\mathbf{C}\|_0}{m}\log(\max\{n_1,n_2,n_3,r\}))$   \\
		 Sparse nonnegative TD \cite{xu2015alternating}          & Tucker rank      & Gaussian       & Yes   & N/A \\
		 Nonnegative TTD  \cite{lee2016nonnegative}   & tensor train rank& Gaussian       & No    & N/A  \\
		Sparse NTF \cite{newman2019non}             & tubal rank $r$   & Gaussian       & No    & N/A   \\
		 Our method                                   & tubal rank $r$   & general noise  & Yes   & $O(\frac{rn_1n_3+\|\mathcal{B}\|_0}{m}\log(\max\{n_1,n_2\}))$   \\ \hline
		\end{tabular}
	\end{center}
\end{table}

The main contributions of this paper are summarized as follows.
(1) Based on tensor-tensor product, a sparse NTF and completion model
from partial and noisy observations
 is proposed under general noise distributions.
(2) The upper bounds of the estimators of the proposed model are established under general noise observations.
Then the upper bounds are specialized to the widely used noise observations including additive Gaussian noise,
additive Laplace noise, and Poisson observations.
(3) The minimax lower bounds  are derived for the previous noise observations,
which match the upper bounds up to a logarithmic factor for different noise models.
(4) An alternating direction method of multipliers (ADMM) based algorithm \cite{Gabay1976A, wang2015global}  is developed to solve the resulting model.
And numerical experiments are presented to demonstrate the effectiveness of the proposed tensor-based method
compared with the matrix-based method in \cite{soni2016noisy}.

The rest of this paper is organized as follows.
Some notation and notions are provided in Section \ref{Prelim}.
We propose a sparse NTF and completion model based on tensor-tensor product
from partial and noisy observations in Section \ref{ProMod},
where the observations are corrupted by  a general class of noise.
In Section \ref{upperbound}, the upper bounds of estimators of the proposed model are established,
which are specialized to three widely used noise models
including additive Gaussian noise, additive Laplace noise, and Poisson observations.
Then the minimax lower bounds are also derived for the previous
observation models in Section \ref{lowerbou}.
An ADMM based algorithm is designed  to solve the resulting model in Section \ref{OptimAlg}.
Numerical experiments are reported to validate
the effectiveness of the proposed method in Section \ref{NumeriExper}.
Finally, the conclusions and future work are given in Section \ref{Conclu}.
All proofs of the theoretical results  are delegated
to the appendix.

\section{Preliminaries}\label{Prelim}

Throughout this paper, $\mathbb{R}$
represents the space with real numbers.
$\mathbb{R}_+^{n_1\times n_2\times n_3}$ denotes the third-order tensor space that
all elements of tensors are nonnegative,
where the order of  a tensor is the number of dimensions, also known as ways or modes \cite{kolda2009tensor}.
Scalars are represented by lowercase letters, e.g., $x$.
Vectors and matrices are represented by
lowercase boldface letters and uppercase boldface letters, respectively,
e.g., $\mathbf{x}$ and $\mathbf{X}$.
Tensors are denoted by capital Euler script letters, e.g., $\mathcal{X}$.
The $(i,j,k)$th entry of a tensor $\mathcal{X}$ is denoted as $\mathcal{X}_{ijk}$.
The $i$th frontal slice of a tensor $\mathcal{X}$ is a matrix denoted by $\mathbf{X}^{(i)}$,
which is a matrix by fixing the third index and varying the first two indexes of $\mathcal{X}$.

The $\ell_2$ norm of a vector $\mathbf{x}\in\mathbb{R}^{n}$,
denoted by $\|\mathbf{x}\|$, is defined as $\|\mathbf{x}\|=\sqrt{\sum_{i=1}^{n}x_i^2}$,
where $x_i$ is the $i$th component of $\mathbf{x}$.
The tensor $\ell_\infty$ norm of a tensor
$\mathcal{X}\in\mathbb{R}^{n_1\times n_2\times n_3}$ is defined as $\|\mathcal{X}\|_\infty=\max_{i,j,k}|\mathcal{X}_{ijk}|$.
The tensor $\ell_0$ norm of $\mathcal{X}$, denoted by $\|\mathcal{X}\|_0$, is defined as the count of all nonzero entries of $\mathcal{X}$.
The inner product of two tensors $\mathcal{X}, \mathcal{Y}\in\mathbb{R}^{n_1\times n_2\times n_3}$
is defined as $\langle \mathcal{X}, \mathcal{Y} \rangle=\sum_{i=1}^{n_3}\langle \mathbf{X}^{(i)}, \mathbf{Y}^{(i)} \rangle$,
where $\langle \mathbf{X}^{(i)}, \mathbf{Y}^{(i)} \rangle=tr((\mathbf{X}^{(i)})^T\mathbf{Y}^{(i)})$.
Here $\cdot^T$ and $tr(\cdot)$   denote the transpose and the  trace of a matrix, respectively.
The tensor Frobenius norm of $\mathcal{X}$ is defined as $\|\mathcal{X}\|_F=\sqrt{\langle \mathcal{X},\mathcal{X} \rangle}$.

Let $p_{x_1}(y_1)$ and $p_{x_2}(y_2)$ be the
probability density functions or  probability mass functions
with respect to the random variables $y_1$ and $y_2$ with parameters $x_1$ and $x_2$, respectively.
The Kullback-Leibler (KL) divergence of $p_{x_1}(y_1)$ from $p_{x_2}(y_2)$ is defined as
$$
D(p_{x_1}(y_1)||p_{x_2}(y_2))=\mathbb{E}_{p_{x_1}(y_1)}\left[\log\frac{p_{x_1}(y_1)}{p_{x_2}(y_2)}\right].
$$
The Hellinger affinity between $p_{x_1}(y_1)$ and $p_{x_2}(y_2)$   is defined as
$$
H(p_{x_1}(y_1)||p_{x_2}(y_2))
=\mathbb{E}_{p_{x_1}}\left[\sqrt{\frac{p_{x_2}(y_2)}{p_{x_1}(y_1)}}\right]
=\mathbb{E}_{p_{x_2}}\left[\sqrt{\frac{p_{x_1}(y_1)}{p_{x_2}(y_2)}}\right].
$$
The joint distributions of higher-order
and multi-dimensional random  variables, denoted
by $p_{\mathcal{X}_1}(\mathcal{Y}), p_{\mathcal{X}_2}(\mathcal{Y})$,
are the joint distributions of the vectorization of tensors.
Then the KL divergence of $p_{\mathcal{X}_1}(\mathcal{Y})$ from $ p_{\mathcal{X}_2}(\mathcal{Y})$ is defined as
$$
D(p_{\mathcal{X}_1}(\mathcal{Y})||p_{\mathcal{X}_2}(\mathcal{Y}))
:=\sum_{i,j,k}D(p_{(\mathcal{X}_1)_{ijk}}(\mathcal{Y}_{ijk})||p_{(\mathcal{X}_2)_{ijk}}(\mathcal{Y}_{ijk})),
$$
and its Hellinger affinity is defined as
$$
H(p_{\mathcal{X}_1}(\mathcal{Y})||p_{\mathcal{X}_2}(\mathcal{Y}))
:=\prod_{i,j,k}H(p_{(\mathcal{X}_1)_{ijk}}(\mathcal{Y}_{ijk}),p_{(\mathcal{X}_2)_{ijk}}(\mathcal{Y}_{ijk})).
$$

Now we define the tensor-tensor product between two third-order tensors \cite{Kilmer2011Factorization}.
\begin{definition}\label{TenTensPro}
	\cite[Definition 3.1]{Kilmer2011Factorization}
Let $\mathcal{X}\in\mathbb{R}^{n_1\times n_2\times n_3}$
and $\mathcal{Y}\in\mathbb{R}^{n_2\times n_4\times n_3}$.
The tensor-tensor product, denoted as  $\mathcal{X}\diamond\mathcal{Y}$,
is an $n_1\times n_4\times n_3$ tensor defined by
$$
\mathcal{X}\diamond\mathcal{Y}:=
\textup{Fold}\left(\textup{Circ}(\textup{Unfold}(\mathcal{X}))\cdot \textup{Unfold}(\mathcal{Y})\right),
$$
where
$$
\textup{Unfold}(\mathcal{X})=\begin{pmatrix} \mathbf{X}^{(1)} \\  \mathbf{X}^{(2)}
\\ \vdots \\  \mathbf{X}^{(n_3)} \end{pmatrix},  \
\textup{Fold}\begin{pmatrix} \mathbf{X}^{(1)} \\  \mathbf{X}^{(2)}
\\ \vdots \\  \mathbf{X}^{(n_3)} \end{pmatrix}=\mathcal{X}, \
\textup{Circ}\begin{pmatrix} \mathbf{X}^{(1)} \\  \mathbf{X}^{(2)}
\\ \vdots \\  \mathbf{X}^{(n_3)} \end{pmatrix}=
\begin{pmatrix}
\mathbf{X}^{(1)}   &  \mathbf{X}^{(n_3)} & \cdots &  \mathbf{X}^{(2)} \\
 \mathbf{X}^{(2)}  & \mathbf{X}^{(1)}  & \cdots &  \mathbf{X}^{(3)}\\
 \vdots            &  \vdots           &        \ddots & \vdots \\
\mathbf{X}^{(n_3)} &\mathbf{X}^{(n_3-1)}&\cdots &  \mathbf{X}^{(1)} \end{pmatrix}.
$$
\end{definition}

By the block circulant structure,
the tensor-tensor product of two third-order tensors
can be implemented efficiently by fast Fourier transform \cite{Kilmer2011Factorization}.
\begin{definition}\cite[Definition 3.14]{Kilmer2011Factorization}
The transpose of a tensor $\mathcal{X}\in\mathbb{R}^{n_1\times n_2\times n_3}$,
is the tensor $\mathcal{X}^T\in\mathbb{R}^{n_2\times n_1\times n_3}$  obtained by transposing each of the frontal
slices and then reversing the order of transposed frontal slices 2 through $n_3$, i.e.,	
$$
(\mathcal{X}^T)^{(1)} = (\mathbf{X}^{(1)})^T,  \  (\mathcal{X}^T)^{(i)} = (\mathbf{X}^{(n_3+2-i)})^T,  \ i=2,\ldots, n_3.
$$
\end{definition}
\begin{definition}\cite[Definition 3.4]{Kilmer2011Factorization}
An $n\times n\times m$ identity tensor $\mathcal{I}$ is the tensor whose
first frontal slice is the $n\times n$ identity matrix, and whose other frontal slices are all zeros.
\end{definition}

\begin{definition}\cite[Definition 3.5]{Kilmer2011Factorization}
A tensor $\mathcal{A}\in\mathbb{R}^{n\times n\times m}$ is said to have an inverse, denoted by $\mathcal{A}^{-1}\in\mathbb{R}^{n\times n\times m}$, if $\mathcal{A}\diamond\mathcal{A}^{-1}=\mathcal{A}^{-1}\diamond\mathcal{A}=\mathcal{I}$, where $\mathcal{I}\in\mathbb{R}^{n\times n\times m}$ is the identity tensor.
\end{definition}

The proximal mapping of a closed proper function $f:\mathfrak{C}\rightarrow (-\infty, +\infty]$ is defined as
$$
\textup{Prox}_{f}(y)=\arg\min_{x\in\mathfrak{C}}\left\{f(x)+\frac{1}{2}\|x-y\|^2\right\},
$$
where $\mathfrak{C}$ is a finite-dimensional Euclidean space
with endowed
inner product $\langle \cdot, \cdot \rangle$ and endowed norm $\|\cdot\|$.
Next we provide a brief summary of the notation used throughout this paper.
\begin{itemize}
\item $\lfloor x\rfloor$ is the integer part of $x$.
$\lceil x\rceil$ is smallest integer that is larger or equal to $x$.
\item Denote $m\vee n=\max\{m,n\}$ and $m\wedge n=\min\{m,n\}$.
\end{itemize}

\section{Sparse NTF and Completion  via Tensor-Tensor Product}\label{ProMod}

Let  $\mathcal{X}^*\in\mathbb{R}_+^{n_1\times n_2\times n_3}$ be an unknown nonnegative tensor we aim to estimate,
which admits a following nonnegative factorization:
$$
\mathcal{X}^*=\mathcal{A}^*	\diamond \mathcal{B}^*,
$$
where $\mathcal{A}^*\in\mathbb{R}_+^{n_1\times r\times n_3}$ and
 $\mathcal{B}^*\in\mathbb{R}_+^{r\times n_2\times n_3}$
 are prior unknown factor tensors with $r\leq \min\{n_1,n_2\}$.
 We assume that each entries of $\mathcal{X}^*, \mathcal{A}^*,
 \mathcal{B}^*$ are bounded, i.e.,
$$
0\leq \mathcal{X}_{ijk}^*\leq \frac{c}{2},  \ \ \
0\leq \mathcal{A}_{ijk}^*\leq 1, \ \ \ 0\leq \mathcal{B}_{ijk}^*\leq b, \ \ \ \forall \ i,j,k,
$$
where $\frac{c}{2}$ is used for simplicity of subsequent analysis.
We remark that the amplitude $1$ of each entry $\mathcal{A}_{ijk}$ of $\mathcal{A}^*$ can be arbitrary.
Besides, our focus is that the factor tensor $\mathcal{B}^*$ is sparse.

However,
only a noisy and incompleted  version of the underlying tensor $\mathcal{X}^*$ is available in practice.
Let $\Omega\subseteq\{1,2,\ldots, n_1\}\times \{1,2,\ldots, n_2\}\times \{1,2,\ldots, n_3\}$
be a subset at which the entries of the observations $\mathcal{Y}$ are collected.
Denote $\mathcal{Y}_{\Omega}\in\mathbb{R}^m$ to be a vector
such that the entries of $\mathcal{Y}$ in the index $\Omega$ are vectorized into a vector by lexicographic order,
where $m$ is the number of observed entries.
Assume that $n_1, n_2, n_3\geq 2$ throughout this paper.
Suppose that the location
 set $\Omega$ is generated according to an independent
Bernoulli model with probability $\gamma=\frac{m}{n_1n_2n_3}$ (denoted by Bern($\gamma$)),
i.e., each index $(i,j,k)$ belongs to $\Omega$ with probability $\gamma$, which is denoted as $\Omega\sim  \text{Bern}(\gamma)$.
Mathematically, the joint probability density function
or probability mass function of the observations $\mathcal{Y}_\Omega$ is given by
\begin{equation}\label{obserPo}
p_{\mathcal{X}_\Omega^*}(\mathcal{Y}_{\Omega})
:=\prod_{(i,j,k)\in \Omega}p_{\mathcal{X}_{ijk}^*}(\mathcal{Y}_{ijk}).
\end{equation}

By  maximum likelihood estimation, we propose the following sparse NTF and completion model with nonnegative constraints:
\begin{equation}\label{model}
\widetilde{\mathcal{X}}^{\lambda}\in\arg\min_{\mathcal{X}=\mathcal{A}	\diamond \mathcal{B}\in\Gamma}\left\{-\log p_{\mathcal{X}_\Omega}(\mathcal{Y}_{\Omega})+\lambda\|\mathcal{B}\|_0\right\},
\end{equation}
where $\lambda>0$ is the regularization parameter and $\Gamma$ is defined by
\begin{equation}\label{TauSet}
\Gamma:=\{\mathcal{X}=\mathcal{A}	\diamond \mathcal{B}:
\ \mathcal{A}\in\mathfrak{L}, \ \mathcal{B}\in\mathfrak{D}, \ 0\leq \mathcal{X}_{ijk}\leq c \}.
\end{equation}
Here $\Gamma$ is a countable set of estimators constructed as follows:
First, let
\begin{equation}\label{denu}
\vartheta:=2^{\lceil\beta\log_2(n_1\vee n_2)\rceil}
\end{equation}
for a specified $\beta\geq 3,$
we construct $\mathfrak{L}$ to be the set
of all tensors $\mathcal{A}\in\mathbb{R}_+^{n_1\times r\times n_3}$
whose entries are discretized to one of $\vartheta$
uniformly sized bins in the range $[0,1]$,
and $\mathfrak{D}$
to be the set of all tensors $\mathcal{B}\in\mathbb{R}_+^{r\times n_2\times n_3}$
whose entries either take the value $0$, or are discretized to
one of $\vartheta$ uniformly sized bins in the range $[0,b]$.

\begin{remark}
When  all entries of $\mathcal{Y}$ are observed and $\mathcal{Y}$ is corrupted by additive Gaussian noise,
the model (\ref{model}) reduces to sparse NTF with tensor-tensor product,
whose relaxation, replaced the tensor $\ell_0$ norm by the tensor $\ell_1$ norm,
 has been applied in patch-based dictionary learning for image data \cite{soltani2016tensor, newman2019non}.
\end{remark}

\begin{remark}
We do not specialize the noise in model (\ref{model}), and just need the joint
 probability density function
or probability mass function of observations in (\ref{obserPo}).
In particular, our model can address the observations with some widely used noise distributions,
such as additive Gaussian noise, additive Laplace noise, and Poisson observations.
\end{remark}

\section{Upper Bounds}\label{upperbound}

In this section, we establish a general upper error
bound of the sparse NTF and completion model from partial observations  under a general class of noise in (\ref{model}),
and then derive the upper bounds of the special noise models
including additive Gaussian noise, additive Laplace noise, and Poisson observations.

Now we establish the upper error bound of the
estimator $\widetilde{\mathcal{X}}^{\lambda}$ in (\ref{model}),
whose proof follows the line of the proof of \cite[Theorem 1]{soni2016noisy},
see also \cite[Theorem 3]{raginsky2010compressed}.
The key technique of this proof is the well-known Kraft-McMillan inequality \cite{Brockway1957Two, kraft1949device}.
And then we construct the penalty of the underlying tensor $\mathcal{X}$ with
the tensor-tensor product of two nonnegative tensors, where one factor tensor is sparse.

\begin{theorem}\label{maintheo}
Suppose that
$\kappa\geq \max_{\mathcal{X}\in\Gamma}\max_{i,j,k} D(p_{\mathcal{X}_{ijk}^*}||p_{\mathcal{X}_{ijk}})$.
Let $\Omega\sim  \textup{Bern}(\gamma)$, where $\gamma=\frac{m}{n_1n_2n_3}$ and $4\leq m\leq n_1n_2n_3$.
Then, for any
$
\lambda\geq 4(\beta+2)\left( 1+\frac{2\kappa}{3}\right) \log(n_1\vee n_2)
$,
the estimator $\widetilde{\mathcal{X}}^{\lambda}$ in (\ref{model}) satisfies
\[
\begin{split}
&~ \frac{\mathbb{E}_{\Omega,\mathcal{Y}_{\Omega}}[-2\log H(p_{\widetilde{\mathcal{X}}^{\lambda}},p_{\mathcal{X}^*})]}{n_1n_2n_3}\\
\leq & \  3\min_{\mathcal{X}=\mathcal{A}\diamond\mathcal{B}\in\Gamma}\left\lbrace
\frac{D(p_{\mathcal{X}^*}|| p_{\mathcal{X}})}{n_1n_2n_3}+
\left( \lambda+\frac{8\kappa(\beta+2) \log(n_1\vee n_2)}{3}\right)
\frac{rn_1n_3+\|\mathcal{B}\|_0}{m} \right\rbrace \\
& \  +\frac{8\kappa\log(m)}{m}.
\end{split}
\]
\end{theorem}

The detailed proof of Theorem \ref{maintheo} is left to Appendix \ref{ProoA}.
From Theorem \ref{maintheo}, we can observe that the
upper bound of $\frac{\mathbb{E}_{\Omega,\mathcal{Y}_{\Omega}}
\left[-2\log H(p_{\widetilde{\mathcal{X}}^\lambda},p_{\mathcal{X}^*})\right]}{n_1n_2n_3}$ is of the order of $O(
\frac{rn_1n_3+\|\mathcal{B}\|_0}{m}\log(n_1\vee n_2))$
if the KL divergence $D(p_{\mathcal{X}^*}|| p_{\mathcal{X}})$
is not too large in the set $\Gamma$.
The explicit upper bounds with respect to $D(p_{\mathcal{X}^*}|| p_{\mathcal{X}})$ in $\Gamma$ and $\kappa$
will be given for the observations with special noise distributions.

\begin{remark}
For the upper error bounds of estimators  of observations with special noise distributions,
the main difference of proofs between the matrix case \cite{soni2016noisy} and the tensor case is to establish the upper bound of $\min_{\mathcal{X}\in\Gamma}
\|\mathcal{X}^*-\mathcal{X}\|_F^2$, where $\Gamma$ is defined as (\ref{TauSet}).
We need to estimate this bound based on the tensor-tensor product structure $\mathcal{X}=\mathcal{A}	\diamond \mathcal{B}\in \Gamma$,
which can be obtained by Lemma \ref{xxappr}.
The key issue in Lemma  \ref{xxappr} is to construct the surrogates of entries
 of the two factor tensors $\mathcal{A}^*, \mathcal{B}^*$  in the set $\Gamma$,
where $\mathcal{X}^*=\mathcal{A}^*\diamond \mathcal{B}^*$.
A main difference between the tensor case in Lemma \ref{xxappr}   and the matrix
case in \cite{soni2016noisy} is the estimation of the upper bounds of  the tensor infinity
norms of $\mathcal{A}^*\diamond\Delta_{\mathcal{B}^*}$ in (\ref{ABDSt}),
$\Delta_{\mathcal{A}^*}\diamond\mathcal{B}^*, \Delta_{\mathcal{A}^*}\diamond\Delta_{\mathcal{B}^*}$,
and the estimation of the upper bound of (17) in Appendix A in \cite{soni2016noisy},
where the block  circulant structure  of the tensor-tensor product is used in (\ref{ABDSt}).
Moreover, the construction of $\beta$ in (\ref{beta}) is also different from the matrix case,
which will influence the estimation of $\vartheta$ in (\ref{vuppb}).
\end{remark}

In the following subsections, we establish the upper error bounds of the estimators for the observations with three special noise models,
 including additive Gaussian noise,
additive Laplace noise, and Poisson observations.
By Theorem \ref{maintheo},
the main steps of proofs for the special noise models are to establish the lower bound of $-2\log H(p_{\widetilde{\mathcal{X}}^{\lambda}},p_{\mathcal{X}^*})$
and the upper bound of $\min_{\mathcal{X}=\mathcal{A}\diamond\mathcal{B}\in\Gamma}D(p_{\mathcal{X}^*}|| p_{\mathcal{X}})$, respectively.

Before deriving the upper error bounds of the  observations with special noise models,
we fix the choices of $\beta$ and $\lambda$ based on Theorem \ref{maintheo}, which are defined as follows:
\begin{equation}\label{beta}
\beta=\max\left\{3,1+\frac{\log(3rn_3^{1.5}b/c)}{\log(n_1\vee n_2)}\right\}
\end{equation}
and
\begin{equation}\label{lambda}
\lambda=4(\beta+2)\left( 1+\frac{2\kappa}{3}\right) \log\left(n_1\vee n_2\right).
\end{equation}

\subsection{Additive Gaussian Noise}

Assume that each entry of the underlying tensor
is corrupted by independently additive zero-mean Gaussian noise with standard deviation $\sigma>0$,
that is
\begin{equation}\label{Gauyom}
\mathcal{Y}_{ijk}=\mathcal{X}_{ijk}^*+\sigma^2\epsilon_{ijk},
\end{equation}
where $\epsilon_{ijk}$ obeys the independently standard normal distribution (i.e., $\epsilon_{ijk}\sim N(0,1)$) for any $(i,j,k)\in\Omega$.
Then the observations $\mathcal{Y}_\Omega$ can be regarded as a vector and
its joint probability density function in (\ref{obserPo}) is given as
\begin{equation}\label{YomeGasu}
p_{\mathcal{X}_\Omega^*}(\mathcal{Y}_{\Omega})
=\frac{1}{(2\pi\sigma^2)^{|\Omega|/2}}\exp\left(-\frac{1}{2\sigma^2}\|\mathcal{Y}_\Omega-\mathcal{X}_\Omega^*\|^2\right),
\end{equation}
where $|\Omega|$ denotes the cardinality of $\Omega$, i.e., $|\Omega|=m$.

Now we establish the explicit upper error bound of the estimator in (\ref{model})
 with the observations $\mathcal{Y}_{\Omega}$ satisfying (\ref{Gauyom}).

\begin{Prop}\label{Gauuupp}
Let $\Omega\sim  \textup{Bern}(\gamma)$, where $\gamma=\frac{m}{n_1n_2n_3}$ and $4\leq m\leq n_1n_2n_3$.
Assume that  $\beta$ and $\lambda$  are defined as (\ref{beta}) and (\ref{lambda}), respectively,
where $\kappa=\frac{c^2}{2\sigma^2}$ in (\ref{lambda}).
Suppose that $\mathcal{Y}_{\Omega}$ satisfies (\ref{Gauyom}).
Then the estimator $\widetilde{\mathcal{X}}^{\lambda}$ in (\ref{model}) satisfies
\[
\begin{split}
\frac{\mathbb{E}_{\Omega,\mathcal{Y}_{\Omega}}
[\|\widetilde{\mathcal{X}}^{\lambda}-\mathcal{X}^*\|_F^2]}{n_1n_2n_3}
\leq  \frac{22c^2\log(m)}{m} + 16(3\sigma^2+2c^2)(\beta+2)
\left(\frac{rn_1n_3+\|\mathcal{B}^*\|_0}{m}\right)\log(n_1\vee n_2).
\end{split}
\]
\end{Prop}

The detailed proof of Proposition \ref{Gauuupp} is left to Appendix \ref{ProoB}.
From Proposition \ref{Gauuupp}, we can see that the upper bound of
$\frac{\mathbb{E}_{\Omega,\mathcal{Y}_{\Omega}}
\left[\|\widetilde{\mathcal{X}}^{\lambda}-\mathcal{X}^*\|_F^2\right]}{n_1n_2n_3}$
for the observations with  additive Gaussian noise
 is of the order  $O(
(\sigma^2+c^2)(\frac{rn_1n_3+\|\mathcal{B}^*\|_0}{m})\log(n_1\vee n_2))$.
Now we give a comparison with a matrix-based method in \cite[Corollary 3]{soni2016noisy}
if we ignore the intrinsic structure of a tensor.
Note that we cannot compare with the matrix-based method directly since the underlying data is the tensor structure.
However,
we can stack these frontal slices of the underlying tensor (with size $n_1\times n_2\times n_3$)
into a matrix, whose size is $n_1n_3\times n_2$.
In this case, the estimator $\mathcal{X}_1$ obtained
by the matrix-based method in  \cite[Corollary 3]{soni2016noisy} satisfies
\begin{equation}\label{MBMGN}
\frac{\mathbb{E}_{\Omega,\mathcal{Y}_{\Omega}}
\left[\|\mathcal{X}_1-\mathcal{X}^*\|_F^2\right]}{n_1n_2n_3}
=  O\left((\sigma^2+c^2)\left(\frac{\widetilde{r}n_1n_3
+\|\mathcal{B}^*\|_0}{m}\right)\log((n_1n_3)\vee n_2)\right),
\end{equation}
where $\widetilde{r}$ is the rank of  the resulting matrix.
In particular, we choose $\widetilde{r}$ in the matrix-based method  the same as $r$ in the tensor-based method with tensor-tensor product.
In real-world applications, $n_1n_3>n_2$ in general.
For example, if $n_3$ denotes the frame
in video datasets or spectral dimensions in hyperspectral image datasets, $n_3$ is large.
Therefore, if $n_1n_3>n_2$, the upper error bound of the matrix-based method in (\ref{MBMGN})
is larger than that of the tensor-based method in Proposition \ref{Gauuupp}.
Especially, when $n_1=n_2$, the logarithmic factor
in Proposition \ref{Gauuupp} is $\log(n_1)$, while it is $\log(n_1n_3)=\log(n_1)+\log(n_3)$
in (\ref{MBMGN}).

 Based on CP decomposition, Jain et al. \cite{jain2017noisy} proposed a tensor
completion method with noisy observations, where the third factor matrix is sparse in  CP decomposition.
For the observations with additive Gaussian noise, the upper error bound of the estimator in \cite[Corollary 1]{jain2017noisy} is of the order
$$
O\left((\sigma^2+c^2)\left(\frac{(n_1+n_2)r'+\|\mathbf{C}\|_0}{m}\right)\log(\max\{n_1,n_2,n_3,r'\})\right),
$$
where $r'$ is the CP rank of the underlying tensor and $\mathbf{C}$ is the third factor matrix in CP decomposition.
This upper bound is hard to compare with other methods since its CP rank is
not comparable with other ranks of tensors in general and the sparse factor is also different.
Moreover, computing the CP rank of a tensor is generally NP-hard \cite{hillar2013most}.

\begin{remark}
We also compare the upper error bound in Proposition \ref{Gauuupp} with that of the noisy tensor completion problem in \cite{wang2019noisy},
which did not consider the sparse factor.
The upper error bound of the estimator $\mathcal{X}_t$ in \cite[Theorem 1]{wang2019noisy} satisfies
\begin{equation}\label{upbtc}
\frac{\|\mathcal{X}_t-\mathcal{X}^*\|_F^2}{n_1n_2n_3}\leq C_t(\sigma^2\vee c^2)\left(\frac{r\max\{n_1,n_2\}n_3}{m}\right)\log((n_1+n_2)n_3)
\end{equation}
with high probability, where $C_t>0$ is a constant.
We note that the upper  error bound of our method can be improved
potentially when $n_2>n_1$ and $\mathcal{B}^*$ is sparse.
In fact, the upper bound in Proposition \ref{Gauuupp} is of the order $O(\frac{rn_1n_3}{m}\log(n_2))$, while the
upper bound in \cite{wang2019noisy} is of the order  $O(\frac{rn_2 n_3}{m}\log((n_1+n_2)n_3))$.
However, when $n_1\geq n_2$, the improvement of the upper bound  of Proposition  \ref{Gauuupp} is mainly on
the logarithmic  factor, which is much smaller than that of (\ref{upbtc}).
Furthermore, the two upper bounds roughly coincide except for the logarithmic factor
when $\mathcal{B}^*$ is not sparse, i.e., $\|\mathcal{B}^*\|_0=rn_2n_3$.
\end{remark}

\begin{remark}
	From Proposition  \ref{Gauuupp}, we know that the upper error bound  decreases when the number of observations increases.
In particular,
when we observe all entries of $\mathcal{Y}$, i.e., $m=n_1n_2n_3$,
 the upper error bound in Proposition  \ref{Gauuupp} is that of the   sparse NTF model
with tensor-tensor product in \cite{newman2019non, soltani2016tensor},
which has been used to construct a tensor patch dictionary prior for CT and facial images, respectively.
This demonstrates that the upper error bound of sparse NTF with tensor-tensor
product in \cite{newman2019non, soltani2016tensor} is lower than that of
sparse NMF in theory, where Soltani et al. \cite{soltani2016tensor} just
showed the performance of sparse NTF with tensor-tensor
product is better than that of sparse NMF in experiments.
\end{remark}

\subsection{Additive Laplace Noise}

Suppose that each entry of the underlying tensor
is corrupted by independently additive Laplace noise with
the location parameter being zero and the diversity being $\tau>0$ (denoted by Laplace($0,\tau$)),
that is
\begin{equation}\label{Lapayom}
\mathcal{Y}_{ijk}=\mathcal{X}_{ijk}^*+\epsilon_{ijk},
\end{equation}
where $\epsilon_{ijk}\sim$ Laplace($0,\tau$)  for any $(i,j,k)\in\Omega$.
Then the joint probability density function of the observations $\mathcal{Y}_\Omega$
is given by
\begin{equation}\label{lapnoid}
p_{\mathcal{X}_{\Omega}^*}(\mathcal{Y}_{\Omega})
=\left(\frac{1}{2\tau}\right)^{|\Omega|}\exp\left(-\frac{\|\mathcal{Y}_{\Omega}-\mathcal{X}_{\Omega}^*\|_1}{\tau}\right).
\end{equation}

Now we establish the upper error bound of the  estimator in (\ref{model})
 for the observations with additive Laplace noise.
\begin{Prop}\label{AddErp}
Let $\Omega\sim  \textup{Bern}(\gamma)$, where $\gamma=\frac{m}{n_1n_2n_3}$ and $4\leq m\leq n_1n_2n_3$.
Assume that $\mathcal{Y}_{\Omega}$ obeys to (\ref{Lapayom}).
Let $\beta$ and  $\lambda$ be defined as (\ref{beta})
and (\ref{lambda}), respectively, where $\kappa= \frac{c^2}{2\tau^2}$ in (\ref{lambda}).
Then the estimator $\widetilde{\mathcal{X}}^{\lambda}$ in (\ref{model}) satisfies
\[
\begin{split}
&~ \frac{\mathbb{E}_{\Omega,\mathcal{Y}_{\Omega}}
[\|\widetilde{\mathcal{X}}^{\lambda}-\mathcal{X}^*\|_F^2]}{n_1n_2n_3}\\
\leq & \ \frac{11c^2(2\tau+c)^2\log(m)}{2m\tau^2}+4\left(3+\frac{2c^2}{\tau^2}\right)(2\tau+c)^2(\beta+2)
\left(\frac{rn_1n_3+\|\mathcal{B}^*\|_0}{m}\right)\log\left(n_1\vee n_2\right).
\end{split}
\]
\end{Prop}

The detailed proof of Proposition \ref{AddErp} is delegated to Appendix \ref{ProoC}.
Similar to the case of the observations with  additive Gaussian noise, we compare
the upper error bound in Proposition \ref{AddErp} with that of \cite[Corollary 5]{soni2016noisy},
which satisfies
\begin{equation}\label{LapMaxm}
\frac{\mathbb{E}_{\Omega,\mathcal{Y}_{\Omega}}
\left[\|\mathcal{X}_2-\mathcal{X}^*\|_F^2\right]}{n_1n_2n_3}=O\left(\frac{(\tau+c)^2c}{\tau} \left(\frac{\widetilde{r}n_1n_3+\|\mathcal{B}^*\|_0}{m}\right)\log((n_1n_3)\vee n_2)\right),
\end{equation}
where $\mathcal{X}_2$ is the estimator by the matrix-based method and
$\widetilde{r}$ is the rank of the resulting matrix by matricizing the underlying tensor.
Therefore, the difference of the upper error bounds
between Proposition \ref{AddErp} and \cite[Corollary 5]{soni2016noisy}
 is mainly on the logarithmic factor.
If $n_1n_3>n_2$, which holds in various real-world scenarios,
the logarithmic factor in (\ref{LapMaxm}) is $\log(n_1n_3)$,
while it is $\log(n_1\vee n_2)$ in Proposition \ref{AddErp}.
In particular, when $n_1=n_2$,
the logarithmic factor in (\ref{LapMaxm}) is $\log(n_1n_3)$,
while it is $\log(n_1)$ in Proposition \ref{AddErp}.

\subsection{Poisson Observations}

Suppose that each entry of $\mathcal{Y}_\Omega$ follows a Poisson distribution,
i.e.,
\begin{equation}\label{Posyijk}
\mathcal{Y}_{ijk}=\text{Poisson}(\mathcal{X}_{ijk}^*),  \ \  \forall \  (i,j,k)\in\Omega,
\end{equation}
where $y=\text{Poisson}(x)$ denotes that $y$ obeys a Poisson distribution
with parameter $x>0$, each $\mathcal{Y}_{ijk}$ is independent and $\mathcal{X}_{ijk}^*>0$.
The joint probability mass function of $\mathcal{Y}_\Omega$ is given as follows:
\begin{equation}\label{Poissobse}
p_{\mathcal{X}_\Omega^*}(\mathcal{Y}_{\Omega})
=\prod_{(i,j,k)\in\Omega}\frac{(\mathcal{X}_{ijk}^*)^{\mathcal{Y}_{ijk}}\exp(-\mathcal{X}_{ijk}^*)}{\mathcal{Y}_{ijk}!}.
\end{equation}

Now we establish the upper error bound of the estimator in (\ref{model}) for the observations obeying (\ref{Posyijk}),
which mainly bases on Theorem \ref{maintheo}.
The key step is to give the upper bound of $D(p_{\mathcal{X}^*}||p_{\mathcal{X}})$.
\begin{Prop}\label{uppPoissobs}
Let $\Omega\sim  \textup{Bern}(\gamma)$, where $\gamma=\frac{m}{n_1n_2n_3}$
and $4\leq m\leq n_1n_2n_3$.
Suppose that each entry of $\mathcal{X}^*$ is positive, i.e.,
$\zeta:=\min_{i,j,k}\mathcal{X}_{ijk}^*>0$,
and each entry of the candidate $\mathcal{X}\in\Gamma$ also satisfies $\mathcal{X}_{ijk}\geq \zeta$.
Let $\beta$ and  $\lambda$  be defined as (\ref{beta})
and (\ref{lambda}), respectively, where $\kappa= {(c-\zeta)}/{\zeta}$ in (\ref{lambda}).
Assume that $\mathcal{Y}_{\Omega}$ obeys to the distribution in (\ref{Posyijk}).
Then the estimator $\widetilde{\mathcal{X}}^{\lambda}$ in (\ref{model}) satisfies
\[
\begin{split}
&~\frac{\mathbb{E}_{\Omega,\mathcal{Y}_{\Omega}}[\|\widetilde{\mathcal{X}}^\lambda-\mathcal{X}^*\|_F^2]}{n_1n_2n_3}\\
\leq & \   \frac{32c(2c-\zeta)^2\log(m)}{\zeta m}+
48c\left(1+\frac{4(c-\zeta)^2}{3\zeta}\right)
\frac{(\beta+2) \left(rn_1n_3+\|\mathcal{B}^*\|_0\right)\log\left(n_1\vee n_2\right)}{m}.
\end{split}
\]
\end{Prop}

We leave the detailed proof of Proposition \ref{uppPoissobs} to Appendix \ref{ProoD}.
Similar to the case of observations with additive Gaussian noise,
we compare the upper error bound in Proposition \ref{uppPoissobs} with
that of the matrix-based method in \cite[Corollary 6]{soni2016noisy}.
The resulting upper error bound of the matrix-based method is of the order
\begin{equation}\label{PoUmbm}
O\left(c\left(1+\frac{c^2}{\zeta}\right)
\left(\frac{\widetilde{r}n_1n_3+\|\mathcal{B}^*\|_0}{m}\right)\log\left((n_1n_3)\vee n_2\right)\right),
\end{equation}
where $\widetilde{r}$ is the rank of the resulting matrix.
The mainly difference of the upper error bounds between the tensor-
and matrix-based methods is the logarithmic factor.
Hence,
if $n_1n_3>n_2$, which holds in various real-world scenarios,
the logarithmic factor in (\ref{PoUmbm}) is $\log(n_1n_3)$,
while it is $\log(n_1\vee n_2)$ in Proposition \ref{uppPoissobs}.
In particular, the  logarithmic factor in Proposition \ref{uppPoissobs} is $\log(n_1)$ when $n_1=n_2$.

\begin{remark}
The constants of the upper bound in  Proposition \ref{uppPoissobs}
have some differences compared with the matrix-based method in \cite[Corollary 6]{soni2016noisy},
which will also influence the recovery error in practice.
\end{remark}

In addition,
Cao et al. \cite{cao2016Poisson} proposed a matrix-based model
for matrix completion with Poisson noise removal
and established the upper error bound of the estimator, where the low-rank
property is utilized by the upper bound of the nuclear norm of a matrix in a constrained set.
The error bound of the estimator $\mathcal{X}_3$ in \cite[Theorem 2]{cao2016Poisson} satisfies
\begin{equation}\label{Poiupbd}
\frac{\|\mathcal{X}_3-\mathcal{X}^*\|_F^2}{n_1n_2n_3}\leq
C_p\left(\frac{c^2\sqrt{\widetilde{r}}}{\zeta}\right)\frac{n_1n_3+n_2}{m}\log^{\frac{3}{2}}(n_1n_2n_3)
\end{equation}
with high probability, where $C_p>0$ is a given constant.
Therefore, if $\log(n_1n_2n_3)>\widetilde{r}$, the upper error bound
of the tensor-based method has a great improvement on the logarithmic factor if $\mathcal{B}^*$ is sparse.
Specifically, when $n_1=n_2$ and $\log(n_1n_2n_3)>\widetilde{r}$,
the logarithmic factor of (\ref{Poiupbd}) is $\log(n_1n_2n_3)$,
while it is $\log(n_1)$ in Proposition \ref{uppPoissobs}.
Recently, Zhang et al. \cite{zhang2021low} proposed a method for low-rank tensor completion with Poisson observations,
which combined the transformed tensor nuclear norm ball constraint with maximum likelihood estimation.
When $m\geq \frac{1}{2}(n_1+n_2)n_3\log(n_1+n_2)$ and all entries of multi-rank of the underlying tensor $\mathcal{X}^*$ are $r_1$,
the upper error bound of the estimator $\mathcal{X}_{tc}$ in \cite[Theorem 3.1]{zhang2021low} is
$$
\frac{\|\mathcal{X}_{tc}-\mathcal{X}^*\|_F^2}{n_1n_2n_3}\leq
C_{tc}n_3\sqrt{\frac{(n_1+n_2)r_1}{m}}\log(n_1n_2n_3)
$$
with high probability, where $C_{tc}>0$ is a given constant. In this case, since $r_1$ is small and  ${(n_1+n_2)r_1}/{m}<1$ in general, the upper error bound in \cite[Theorem 3.1]{zhang2021low} is larger than that in Proposition \ref{uppPoissobs}.

\section{Minimax Lower Bounds}\label{lowerbou}

In this section, we study the sparse NTF and completion problem with incomplete and noisy observations,
and establish the lower bounds on the
minimax risk for the candidate estimator in the following set:
\begin{equation}\label{Ucnae}
\begin{split}
\mathfrak{U}(r,b,s):
=\Big\{\mathcal{X}=\mathcal{A}\diamond\mathcal{B}\in\mathbb{R}_+^{n_1\times n_2\times n_3}:& \
\mathcal{A}\in\mathbb{R}_+^{n_1\times r \times n_3}, \ 0\leq \mathcal{A}_{ijk}\leq 1,\\
& ~ ~ \mathcal{B}\in\mathbb{R}_+^{r\times n_2\times n_3}, \ 0\leq \mathcal{B}_{ijk}\leq b, \ \|\mathcal{B}\|_0\leq s\Big\},
\end{split}
\end{equation}
which implies that the underlying tensor has a nonnegative factorization with tensor-tensor product and one factor tensor is sparse.
We only know the joint probability density function
or probability mass function of observations $\mathcal{Y}_\Omega$ given by (\ref{obserPo}).
Let $\widetilde{\mathcal{X}}$ be an estimator of $\mathcal{X}^*$.
The risk of estimators with incomplete observations is defined as
\begin{equation}\label{mirisk}
\mathfrak{R}_{\widetilde{\mathcal{X}}}
=\frac{\mathbb{E}_{\Omega,\mathcal{Y}_{\Omega}}[\|\widetilde{\mathcal{X}}-\mathcal{X}^*\|_F^2]}{n_1n_2n_3}.
\end{equation}
The worst-case performance of an estimator $\widetilde{\mathcal{X}}$
over the set $\mathfrak{U}(r,b,s)$ is defined as
$$
\inf_{\widetilde{\mathcal{X}}}\sup_{\mathcal{X}^*\in\mathfrak{U}(r,b,s)}\mathfrak{R}_{\widetilde{\mathcal{X}}}.
$$
The estimator is defined to achieve the minimax risk when it is the smallest maximum risk among all possible estimators.
Denote
\begin{equation}\label{deltasp}
\Delta:=\min\left\{1,\frac{s}{n_2n_3}\right\}.
\end{equation}

Now we establish the lower bounds of the minimax risk,
whose proof follows a  similar line of \cite[Theorem 1]{sambasivan2018minimax} for noisy matrix completion,
see also \cite[Theorem 3]{klopp2017robust}.
The main technique is to define suitable packing sets
for two factor tensors $\mathcal{A}$ and $\mathcal{B}$ in (\ref{Ucnae})  based on tensor-tensor product.
Then we construct binary sets for the two packing sets with the tensor structure,
which are subsets of (\ref{Ucnae}).
The line is mainly on the general results for the
risk estimate based on KL divergence \cite[Theorem 2.5]{tsybakov2009}
and the measures of two probability distributions.
In this case, we need to establish the lower bounds of Hamming distance between any two binary
sequences based on Varshamov-Gilbert bound \cite[Lemma 2.9]{tsybakov2009}.

First we establish the minimax lower bound with a general class of noise models in (\ref{obserPo}),
whose joint probability density function
or probability mass function of observations is given.

\begin{theorem}\label{lowbounMai}
Suppose that the KL divergence of the scalar probability density function or probability mass function satisfies
\begin{equation}\label{DKLpq}
D(p(x)||q(x))\leq \frac{1}{2\nu^2}(x-y)^2,
\end{equation}
where $\nu>0$ depends on the distribution of observations in (\ref{obserPo}).
Assume that $\mathcal{Y}_\Omega$ follows from (\ref{obserPo}).
Let $r\leq\min\{n_1,n_2\}$ and
$r\leq s\leq rn_2n_3$.
Then there exist $C,\beta_c>0$ such that the minimax risk in (\ref{mirisk}) satisfies
$$
\inf_{\widetilde{\mathcal{X}}}\sup_{\mathcal{X}^*
\in\mathfrak{U}(r,b,s)}\frac{\mathbb{E}_{\Omega,\mathcal{Y}_\Omega}[\|\widetilde{\mathcal{X}}-\mathcal{X}^*\|_F^2]}{n_1n_2n_3}\geq
C \min\left\{\Delta b^2,\beta_c^2\nu^2\left(\frac{s + rn_1n_3}{m}\right)\right\},
$$
where $\Delta$ is defined as (\ref{deltasp}).
\end{theorem}

From Theorem \ref{lowbounMai}, we know that the minimax lower bound matches
the upper error  bound in Theorem \ref{maintheo} with a logarithmic factor $\log(n_1\vee n_2)$, which
implies that the upper error bound in Theorem \ref{maintheo} is nearly optimal up to
a logarithmic factor $\log(n_1\vee n_2)$.

\begin{remark}
 For the  minimax lower bound with general noise observations in Theorem \ref{lowbounMai}, the main differences of proofs between \cite{sambasivan2018minimax} and Theorem \ref{lowbounMai}
are the constructions of packing sets (the sets in (\ref{GenesubX}), (\ref{CXZG}), (\ref{GenesubXB})) for the set $\mathfrak{U}(r,b,s)$ in (\ref{Ucnae}),
where the tensor-tensor product is used in the set (\ref{GenesubX}).
Moreover, being different from the proof of \cite{sambasivan2018minimax},
we need to construct the subsets of the packing sets
(the sets in (\ref{SubXAA}) and (\ref{SubXBB})),
where  the  tensor in the subsets has special nonnegative tensor factorization structures with the tensor-tensor product form.
The special block tensors are constructed for one factor tensor and special sets with block structure  tensors are
constructed for the other factor tensor (see (\ref{GeneXACsub}) and (\ref{GeneXbSubBb})).
\end{remark}

In the next subsections, we establish the explicit
lower bounds for the special noise distributions,
including additive Gaussian noise, additive Laplace noise, and Poisson observations,
where the condition (\ref{DKLpq}) can be satisfied easily in each case.

\subsection{Additive Gaussian Noise}

In this subsection, we establish the minimax lower bound for  the observations with additive Gaussian noise,
i.e.,  $\mathcal{Y}_\Omega$ obeys to (\ref{Gauyom}).
By Theorem \ref{lowbounMai}, the key issue is to give the explicit  $\nu$ in (\ref{DKLpq}).

\begin{Prop}\label{ProupbG}
Assume that $\mathcal{Y}_\Omega$ follows from (\ref{Gauyom}).
Let $r\leq\min\{n_1,n_2\}$ and
$r\leq s\leq rn_2n_3$.
Then there exist $C,\beta_c>0$ such that the minimax risk in (\ref{mirisk}) satisfies
$$
\inf_{\widetilde{\mathcal{X}}}\sup_{\mathcal{X}^*
\in\mathfrak{U}(r,b,s)}\frac{\mathbb{E}_{\Omega,\mathcal{Y}_\Omega}[\|\widetilde{\mathcal{X}}-\mathcal{X}^*\|_F^2]}{n_1n_2n_3}\geq
C \min\left\{\Delta b^2,\beta_c^2\sigma^2\left(\frac{s + rn_1n_3}{m}\right)\right\},
$$
where $\Delta$ is defined as (\ref{deltasp}).
\end{Prop}

\begin{remark}
From Proposition \ref{ProupbG}, we know that the minmax lower bound matches the upper  error bound
in Proposition \ref{Gauuupp}
up to a  logarithmic factor $\log(n_1\vee n_2)$, which implies that the upper error bound
in Proposition \ref{Gauuupp}  is nearly optimal.
\end{remark}

\begin{remark}
When we observe all entries of $\mathcal{Y}$, i.e., $m=n_1n_2n_3$,   the minimax lower bound  in Proposition \ref{ProupbG}
is just that of sparse NTF with tensor-tensor product, which has been applied in dictionary learning \cite{newman2019non}.
\end{remark}

\subsection{Additive Laplace Noise}

In this subsection, we establish the minimax lower bound for  the observations with additive Laplace noise,
i.e.,  $\mathcal{Y}_\Omega$ obeys to (\ref{Lapayom}).
Similar to the case of  additive Gaussian noise,
we only need to give $\nu$ explicitly in (\ref{DKLpq}) in  Theorem  \ref{lowbounMai}.

\begin{Prop}\label{lapUpb}
Assume that $\mathcal{Y}_\Omega$ follows from (\ref{Lapayom}).
Let $r\leq\min\{n_1,n_2\}$ and
$r\leq s\leq rn_2n_3$.
Then there exist $C,\beta_c>0$ such that the minimax risk in (\ref{mirisk}) satisfies
$$
\inf_{\widetilde{\mathcal{X}}}\sup_{\mathcal{X}^*
\in\mathfrak{U}(r,b,s)}\frac{\mathbb{E}_{\Omega,\mathcal{Y}_\Omega}[\|\widetilde{\mathcal{X}}-\mathcal{X}^*\|_F^2]}{n_1n_2n_3}\geq
C \min\left\{\Delta b^2,\beta_c^2\tau^2\left(\frac{s + rn_1n_3}{m}\right)\right\}.
$$
\end{Prop}

\begin{remark}
It follows from Proposition \ref{lapUpb} that the rate attained
by our estimator in Proposition \ref{AddErp} is optimal up to a logarithmic factor $\log(n_1\vee n_2)$,
which is similar to the case of the observations with additive Gaussian noise.
\end{remark}

\subsection{Poisson Observations}

In this subsection, we establish the minimax lower bound for Poisson observations,
i.e., $\mathcal{Y}_\Omega$ obeys to (\ref{Posyijk}).
There is a slight difference compared with additive Gaussian noise and Laplace noise,
we need to assume that all entries of the underlying tensor are strictly positive,
i.e., $\zeta:=\min_{i,j,k}\mathcal{X}_{ijk}^*>0$.
Suppose that $\zeta<b$.
Being different from the candidate set (\ref{Ucnae}),
each entry of the candidate tensor is also strictly positive.
The candidate set is defined as follows:
\begin{equation}\label{UbarPoissn}
\begin{split}
\widetilde{\mathfrak{U}}(r,b,s,\zeta):
=\Big\{\mathcal{X}=\mathcal{A}\diamond\mathcal{B}\in\mathbb{R}_+^{n_1\times n_2\times n_3}:& \
\mathcal{X}_{ijk}\geq \zeta, \  \mathcal{A}\in\mathbb{R}_+^{n_1\times r \times n_3}, \ 0\leq \mathcal{A}_{ijk}\leq 1,\\
&~ ~  \mathcal{B}\in\mathbb{R}_+^{r\times n_2\times n_3}, \ 0\leq \mathcal{B}_{ijk}\leq b, \ \|\mathcal{B}\|_0\leq s\Big\}.
\end{split}
\end{equation}
Then we know that $\widetilde{\mathfrak{U}}(r,b,s,\zeta)\subseteq\mathfrak{U}(r,b,s)$.

Now the lower bound of candidate estimators for Poisson observations
is given in the following proposition, whose proof
follows a similar line of the matrix case in \cite[Theorem 6]{sambasivan2018minimax}.
For the sake of completeness, we give it here.
Similar to Theorem \ref{lowbounMai}, the main differences
between the matrix- and tensor-based methods are constructions of
the packing sets for the two nonnegative factors $\mathcal{A}$ and $\mathcal{B}$.
We mainly use the results in \cite[Theorem 2.5]{tsybakov2009} for the constructed packing sets
and the Varshamov-Gilbert bound \cite[Lemma 2.9]{tsybakov2009} for the binary sets.

\begin{Prop}\label{Poisslow}
Suppose that  $\mathcal{Y}_\Omega$  follows from (\ref{Posyijk}).
Assume that  $\zeta<b$, where  $\zeta:=\min_{i,j,k}\mathcal{X}_{ijk}^*>0$.
Let $r\leq\min\{n_1,n_2\}$ and
$n_2n_3< s\leq rn_2n_3$.
Then there exist $0<\widetilde{\beta}_c<1$ and $\widetilde{C}>0$ such that
$$
\inf_{\widetilde{\mathcal{X}}}\sup_{\mathcal{X}^*
\in\widetilde{\mathfrak{U}}(r,b,s,\zeta)}\frac{\mathbb{E}_{\Omega,\mathcal{Y}_\Omega}[\|\widetilde{\mathcal{X}}-\mathcal{X}^*\|_F^2]}{n_1n_2n_3}\geq
\widetilde{C}\min\left\{\widetilde{\Delta} b^2,\widetilde{\beta}_c^2\zeta\left(\frac{s-n_2n_3+rn_1n_3}{m}\right)\right\},
$$
where $\widetilde{\Delta}:=\min\{(1-\varsigma)^2, \Delta_1\}$
with $\varsigma:=\frac{\zeta}{b}$ and $\Delta_1:=\min\{1,\frac{s-n_2n_3}{n_2n_3}\}$.
\end{Prop}

\begin{remark}
From Proposition \ref{Poisslow},
we note that the lower bound of Poisson observations is of the order $O(\frac{s-n_2n_3+rn_1n_3}{m})$.
In particular, when $s\geq 2n_2n_3$, the lower bound in Proposition \ref{Poisslow}
matches the upper bound in Proposition \ref{uppPoissobs} up to a logarithmic factor $\log(n_1\vee n_2)$.
\end{remark}

\begin{remark}
For the  minimax lower bound with Poisson observations in Proposition \ref{Poisslow}, the main differences of proofs between \cite{sambasivan2018minimax} and Proposition \ref{Poisslow}
are the constructions of packing sets (the sets in (\ref{PoscsubX1}), (\ref{CXZ}), (\ref{PoscsubX1B1})) for the set $\widetilde{\mathfrak{U}}(r,b,s,\zeta)$ in (\ref{UbarPoissn}),
where the tensor-tensor product is used in the set (\ref{PoscsubX1}).
Moreover, the subsets of the packing sets  with two nonnegative factor tensors
(the sets in (\ref{PoissXA1A}) and (\ref{PoissXB1B})) need to be constructed,  where  the tensor-tensor product is also used in the two subsets.
Besides,  in the two subsets,
the special block tensors  for one factor tensor and special sets with block tensors for the other factor tensor (see the sets in  (\ref{PoisXAsubC1})   and (\ref{PoisXBsubB1}))  are constructed.
\end{remark}

\section{Optimization Algorithm}\label{OptimAlg}

In this section, we present an ADMM based algorithm \cite{Gabay1976A, wang2015global} to solve model (\ref{model}).
Note that the feasible set $\Gamma$ in (\ref{TauSet}) is discrete which makes the algorithm design difficult.
In order to use continuous optimization
techniques, the discrete assumption of $\Gamma$ is dropped.
This may be justified by choosing a very large
value of $\vartheta$ and by noting that continuous optimization algorithms
 use finite precision arithmetic when executed on a computer.
 Now we consider to solve the following relaxation model:
\begin{equation}\label{MidelSolv}
\begin{split}
\min_{\mathcal{X},\mathcal{A},\mathcal{B}} \ & -\log p_{\mathcal{X}_\Omega}(\mathcal{Y}_{\Omega})+\lambda\|\mathcal{B}\|_0 \\
\text{s.t.}\ &  \mathcal{X}=\mathcal{A}\diamond \mathcal{B},\
0\leq \mathcal{X}_{ijk}\leq c,  \
0\leq \mathcal{A}_{ijk}\leq 1,  \
0\leq \mathcal{B}_{ijk}\leq b.
\end{split}
\end{equation}
Let $\mathfrak{X}'=\{\mathcal{X}\in\mathbb{R}_+^{n_1\times n_2\times n_3}:0\leq \mathcal{X}_{ijk}\leq c\}$,
$\mathfrak{A}=\{\mathcal{A}\in\mathbb{R}_+^{n_1\times r\times n_3}:0\leq \mathcal{A}_{ijk}\leq 1\}$,
$\mathfrak{B}'=\{\mathcal{B}\in\mathbb{R}_+^{r\times n_2\times n_3}:0\leq \mathcal{B}_{ijk}\leq b\}$,
and $\mathcal{Q}=\mathcal{X}, \mathcal{M}=\mathcal{A}$, $\mathcal{N}=\mathcal{B}, \mathcal{Z}=\mathcal{B}$.
Then problem (\ref{MidelSolv}) can be rewritten equivalently as
\begin{equation}\label{ModelOtheFo}
\begin{split}
\min_{\mathcal{X},\mathcal{A},\mathcal{B},\mathcal{Q},\mathcal{M},\mathcal{N},\mathcal{Z}} \ &
-\log p_{\mathcal{X}_\Omega}(\mathcal{Y}_{\Omega})+\lambda\|\mathcal{N}\|_0
+\delta_{\mathfrak{X}'}(\mathcal{Q})+
\delta_{\mathfrak{A}}(\mathcal{M})+\delta_{\mathfrak{B}'}(\mathcal{Z}) \\
\text{s.t.}\ &  \mathcal{X}=\mathcal{A}\diamond \mathcal{B},
\mathcal{Q} = \mathcal{X}, \mathcal{M}=\mathcal{A}, \mathcal{N}=\mathcal{B}, \mathcal{Z}=\mathcal{B},
\end{split}
\end{equation}
where $\delta_{\mathfrak{A}}(x)$ denotes the indicator function
of $\mathfrak{A}$, i.e., $\delta_{\mathfrak{A}}(x)=0$ if $x\in \mathfrak{A}$ otherwise $\infty$.
The augmented Lagrangian function associated with (\ref{ModelOtheFo}) is defined as
\[
\begin{split}
&L(\mathcal{X},\mathcal{A},\mathcal{B},\mathcal{Q},\mathcal{M},\mathcal{N},\mathcal{Z},\mathcal{T}_i)\\
:=&
-\log p_{\mathcal{X}_\Omega}(\mathcal{Y}_{\Omega})+\lambda\|\mathcal{N}\|_0
+\delta_{\mathfrak{X}'}(\mathcal{Q})+
\delta_{\mathfrak{A}}(\mathcal{M})+\delta_{\mathfrak{B}'}(\mathcal{Z})-\langle \mathcal{T}_1, \mathcal{X}-\mathcal{A}\diamond \mathcal{B} \rangle\\
&-\langle \mathcal{T}_2, \mathcal{Q}- \mathcal{X}\rangle - \langle\mathcal{T}_3, \mathcal{M}-\mathcal{A} \rangle
-\langle \mathcal{T}_4,  \mathcal{N}-\mathcal{B} \rangle -\langle \mathcal{T}_5, \mathcal{Z}-\mathcal{B}\rangle \\ &+\frac{\rho}{2}\Big(\|\mathcal{X}-\mathcal{A}\diamond \mathcal{B}\|_F^2
+\|\mathcal{Q} - \mathcal{X}\|_F^2+\| \mathcal{M}-\mathcal{A}\|_F^2+\| \mathcal{N}-\mathcal{B}\|_F^2+\|\mathcal{Z}-\mathcal{B}\|_F^2\Big),
\end{split}
\]
where $\mathcal{T}_i$ are the Lagrangian multipliers, $i=1,\ldots, 5$, and
$\rho>0$ is the penalty parameter. The iteration of ADMM is given as follows:
\begin{align}
&\mathcal{X}^{k+1}=\arg\min_{\mathcal{X}} L(\mathcal{X},\mathcal{A}^k,\mathcal{B}^k,\mathcal{Q}^k,\mathcal{M}^k,\mathcal{N}^k,\mathcal{Z}^k,\mathcal{T}_i^k) \nonumber \\ \label{Xk1}
&~~~~~~=\textup{Prox}_{(-\frac{1}{2\rho}\log p_{\mathcal{X}_\Omega}(\mathcal{Y}_{\Omega}))}\left(\frac{1}{2}\left(\mathcal{Q}^k+\mathcal{A}^k\diamond \mathcal{B}^k+\frac{1}{\rho}(\mathcal{T}_1^k-\mathcal{T}_2^k)\right)\right), \\
&\mathcal{A}^{k+1}=\arg\min_{\mathcal{A}} L(\mathcal{X}^{k+1},\mathcal{A},\mathcal{B}^k,\mathcal{Q}^k,\mathcal{M}^k,\mathcal{N}^k,\mathcal{Z}^k,\mathcal{T}_i^k)\nonumber \\ \label{Ak1}
&~~~~~~=\left(\mathcal{M}^k+(\mathcal{X}^{k+1}-\frac{1}{\rho}\mathcal{T}_1^k)\diamond
(\mathcal{B}^k)^T-\frac{1}{\rho}\mathcal{T}_3^k\right)\diamond (\mathcal{B}^k\diamond (\mathcal{B}^k)^T+\mathcal{I})^{-1}, \\
&\mathcal{B}^{k+1}=\arg\min_{\mathcal{B}} L(\mathcal{X}^{k+1},\mathcal{A}^{k+1},\mathcal{B},\mathcal{Q}^k,\mathcal{M}^k,
\mathcal{N}^k,\mathcal{Z}^k,\mathcal{T}_i^k)\nonumber\\ \label{bK1}
&~~~~~~=\left((\mathcal{A}^{k+1})^T\diamond \mathcal{A}^{k+1}+2\mathcal{I}\right)^{-1}\diamond \nonumber  \\
&~~~~~~~~~~
\left((\mathcal{A}^{k+1})^T\diamond\mathcal{X}^{k+1}+\mathcal{N}^k+\mathcal{Z}^k
-\frac{1}{\rho}\left((\mathcal{A}^{k+1})^T\diamond\mathcal{T}_1^k+
\mathcal{T}_4^k+\mathcal{T}_5^k\right)\right), \\
\label{QK1}
&\mathcal{Q}^{k+1}=\arg\min_{\mathcal{Q}} L(\mathcal{X}^{k+1},\mathcal{A}^{k+1},\mathcal{B}^{k+1},\mathcal{Q},\mathcal{M}^{k},\mathcal{N}^k,
\mathcal{Z}^{k},\mathcal{T}_i^k)
=\Pi_{\mathfrak{X}'}\Big(\mathcal{X}^{k+1}+\frac{1}{\rho}\mathcal{T}_2^k\Big), \\ \label{Mk1}
&\mathcal{M}^{k+1}=\arg\min_{\mathcal{M}} L(\mathcal{X}^{k+1},\mathcal{A}^{k+1},\mathcal{B}^{k+1},\mathcal{Q}^{k+1},\mathcal{M},\mathcal{N}^k,
\mathcal{Z}^{k+1},\mathcal{T}_i^k)
=\Pi_{\mathfrak{A}}\Big(\mathcal{A}^{k+1}+\frac{1}{\rho}\mathcal{T}_3^k\Big),\\ \label{Nk1}
&\mathcal{N}^{k+1}=\arg\min_{\mathcal{N}} L(\mathcal{X}^{k+1},\mathcal{A}^{k+1},\mathcal{B}^{k+1},\mathcal{Q}^{k+1},\mathcal{M}^{k+1},\mathcal{N},
\mathcal{Z}^{k},\mathcal{T}_i^k)\nonumber \\
&~~~~~~~=\textup{Prox}_{\frac{\lambda}{\rho}\|\cdot\|_0}\Big(\mathcal{B}^{k+1}+\frac{1}{\rho}\mathcal{T}_4^k\Big), \\ \label{Zk1}
&\mathcal{Z}^{k+1}=\arg\min_{\mathcal{Z}} L(\mathcal{X}^{k+1},\mathcal{A}^{k+1},\mathcal{B}^{k+1},\mathcal{Q}^{k+1},\mathcal{M}^{k+1},
\mathcal{N}^{k+1},\mathcal{Z},\mathcal{T}_i^k)
=\Pi_{\mathfrak{B}'}\Big(\mathcal{B}^{k+1}+\frac{1}{\rho}\mathcal{T}_5^k\Big), \\ \label{Tk12}
&\mathcal{T}_1^{k+1}=\mathcal{T}_1^k-\rho(\mathcal{X}^{k+1}-\mathcal{A}^{k+1}\diamond \mathcal{B}^{k+1}), \
 \mathcal{T}_2^{k+1}=  \mathcal{T}_2^{k}-\rho(\mathcal{Q}^{k+1}- \mathcal{X}^{k+1}), \\  \label{Tk34}
& \mathcal{T}_3^{k+1}=  \mathcal{T}_3^k-\rho(\mathcal{M}^{k+1}-\mathcal{A}^{k+1}), \
\mathcal{T}_4^{k+1}=\mathcal{T}_4^{k}-\rho(\mathcal{N}^{k+1}-\mathcal{B}^{k+1}), \\ \label{Tk5}
&\mathcal{T}_5^{k+1}=\mathcal{T}_5^k-\rho(\mathcal{Z}^{k+1}-\mathcal{B}^{k+1}),
\end{align}
where $\Pi_{\mathfrak{X}'}(\mathcal{X}), \Pi_{\mathfrak{A}}(\mathcal{X})$, and
$\Pi_{\mathfrak{B}'}(\mathcal{X})$ denote
the projections of $\mathcal{X}$ onto the sets $\mathfrak{X}'$, $\mathfrak{A}$, and $\mathfrak{B}'$, respectively.

Now the ADMM for solving  (\ref{ModelOtheFo}) is stated in Algorithm \ref{AlSNDM}.
\begin{algorithm}[htbp]
	\caption{Alternating Direction Method of Multipliers for Solving (\ref{ModelOtheFo})} \label{AlSNDM}
{\bf Input}. Let  $\rho>0$ be a given constant. Given $\mathcal{A}^0, \mathcal{B}^0, \mathcal{Q}^0,
\mathcal{M}^0, \mathcal{N}^0, \mathcal{Z}^0, \mathcal{T}_i^0, i=1,\ldots,5$.
For $k=0,1,\ldots,$ perform the following steps: \\
			{\bf Step 1}.  Compute $\mathcal{X}^{k+1}$  via (\ref{Xk1}). \\
			{\bf Step 2}. Compute $\mathcal{A}^{k+1}$ by (\ref{Ak1}). \\
			{\bf Step 3}.  Compute $\mathcal{B}^{k+1}$ by (\ref{bK1}). \\
	    	{\bf Step 4}. Compute $\mathcal{Q}^{k+1}, \mathcal{M}^{k+1},
\mathcal{N}^{k+1}, \mathcal{Z}^{k+1}$ by (\ref{QK1}), (\ref{Mk1}), (\ref{Nk1}), and (\ref{Zk1}), respectively.  \\
		   {\bf Step 5}. Update  $\mathcal{T}_1^{k+1}$, $\mathcal{T}_2^{k+1}$, $\mathcal{T}_3^{k+1}$,
 $\mathcal{T}_4^{k+1}$, $\mathcal{T}_5^{k+1}$ via  (\ref{Tk12}), (\ref{Tk34}), and (\ref{Tk5}), respectively. \\
			{\bf Step 6}. If a termination criterion is not satisfied, set $k:=k+1$ and go to Step 1.
\end{algorithm}

Algorithm \ref{AlSNDM} is an ADMM based algorithm for solving nonconvex optimization problems.
Although great efforts have been made  about the convergence of ADMM
for nonconvex models in recent years \cite{wang2015global, Hong2016Convergence},
the existing ADMM based algorithm cannot been applied to our model directly
since both the objective function and constraints are nonconvex.
Moreover, the data-fitting term is nonsmooth when the observations are corrupted by additive Laplace noise,
which also gives rise to the difficulty of analyzing the convergence of ADMM.

\begin{remark}\label{theProMap}
In Algorithm \ref{AlSNDM}, one needs to compute the proximal mapping
$\textup{Prox}_{(-\frac{1}{2\rho}\log p_{\mathcal{X}_\Omega}(\mathcal{Y}_{\Omega}))}(\mathcal{S})$,
where $\mathcal{S}=\frac{1}{2}(\mathcal{Q}^k+\mathcal{A}^k\diamond \mathcal{B}^k+\frac{1}{\rho}(\mathcal{T}_1^k-\mathcal{T}_2^k))$.
In particular, for additive Gaussian noise, additive Laplace noise,
and Poisson observations, the proximal mappings at $\mathcal{S}$ are given by
\begin{itemize}
\item Additive Gaussian noise:
$$
\textup{Prox}_{(-\frac{1}{2\rho}\log p_{\mathcal{X}_\Omega}(\mathcal{Y}_{\Omega}))}(\mathcal{S})
=\mathcal{P}_\Omega\left(\frac{\mathcal{Y}+2\rho\sigma^2\mathcal{S}}{1+2\rho\sigma^2}\right)+
\mathcal{P}_{\overline{\Omega}}(\mathcal{S}),
$$
where $\overline{\Omega}$ is the complementary set of $\Omega$.
\item Additive Laplace noise:
$$\textup{Prox}_{(-\frac{1}{2\rho}\log p_{\mathcal{X}_\Omega}(\mathcal{Y}_{\Omega}))}(\mathcal{S})
    =\mathcal{P}_\Omega\left(\mathcal{Y}_{\Omega}+\textup{sign}
    (\mathcal{S}-\mathcal{Y}_{\Omega})\circ\max\left\{|\mathcal{S}
    -\mathcal{Y}_\Omega|-\frac{1}{2\rho\tau},0\right\}\right)+\mathcal{P}_{\overline{\Omega}}(\mathcal{S}),
$$
where $\textup{sign}(\cdot)$ denotes the signum function and $\circ$ denotes the point-wise product.
\item Poisson observations:
$$
\textup{Prox}_{(-\frac{1}{2\rho}\log p_{\mathcal{X}_\Omega}(\mathcal{Y}_{\Omega}))}(\mathcal{S})
    =\mathcal{P}_\Omega\left(\frac{2\rho \mathcal{S}-\mathbb{I}_{n_1n_2}+\sqrt{(2\rho \mathcal{S}-\mathbb{I}_{n_1n_2})^2
    +8\rho \mathcal{Y}}}{4\rho}\right)+\mathcal{P}_{\overline{\Omega}}(\mathcal{S}),
$$
    where $\mathbb{I}_{n_1n_2}\in\mathbb{R}^{n_1\times n_2 \times n_3}$ denotes the tensor with all entries being $1$, and the square and root are performed in point-wise manners.
\end{itemize}
\end{remark}

\begin{remark}
We also need to compute the proximal mapping of the tensor $\ell_0$ norm \cite{donoho1994ideal} in Algorithm \ref{AlSNDM}.
Note that the tensor $\ell_0$ norm is separable.
Therefore, we just need to derive its scalar form.
For any $t>0$, the proximal mapping of $t \|\cdot\|_0$ at $y$ is
given by (see, e.g., \cite[Example 6.10]{beck2017first})
$$
\textup{Prox}_{t\|\cdot\|_0}(y)=
\left\{
\begin{array}{ll}
0, & \mbox{if}  \ |y|<\sqrt{2t}, \\
\{0,y\}, &  \mbox{if} \  |y|=\sqrt{2t}, \\
y, & \mbox{if}\  |y|>\sqrt{2t}.
\end{array}
\right.
$$
\end{remark}

\begin{remark}
	The ADMM based algorithm is developed to solve the model (\ref{ModelOtheFo}). However, the problem
	 (\ref{ModelOtheFo})  is nonconvex, and it is difficult to obtain its globally optimal solution in  experiments,
	 while  the estimators of the  upper error bounds  in Section \ref{upperbound} are globally optimal.
\end{remark}

\begin{remark}
The main cost of ADMM in Algorithm \ref{AlSNDM} is the tensor-tensor product and tensor inverse operations.
First, we consider the computational cost of the tensor-tensor product for two tensors $\mathcal{A}\in \mathbb{R}^{n_1\times r\times n_3}$ and $ \mathcal{B}\in \mathbb{R}^{r\times n_2\times n_3}$,
which is implemented by fast Fourier transform \cite{Kilmer2011Factorization}.
The application of discrete Fourier transform
to an $n_3$-vector is of $O(n_3\log(n_3))$ operations.
After Fourier transform along the tubes, we need to compute $n_3$ matrix products with sizes $n_1$-by-$r$ and $r$-by-$n_2$,
whose cost is $O(rn_1n_2n_3)$. Therefore, for the tensor-tensor product of $\mathcal{A}\in \mathbb{R}^{n_1\times r\times n_3}$ and $ \mathcal{B}\in \mathbb{R}^{r\times n_2\times n_3}$,
the total cost is $O(r(n_1+n_2)n_3\log(n_3)+rn_1n_2n_3)$.
Second, for the inverse operation of an $n\times n\times n_3$ tensor, one takes fast Fourier transform along the third-dimension and operates the inverse for each frontal slice
in the Fourier domain. Then the total cost of the tensor inverse operation is $O(n^2n_3\log(n_3)+n^3n_3)$.
For the ADMM, its main cost is to compute $\mathcal{A}^{k+1}$ and $\mathcal{B}^{k+1}$. The  complexities of computing $\mathcal{A}^{k+1}$ and $\mathcal{B}^{k+1}$ are $O(n_2(r+n_1)n_3\log(n_3)+rn_1n_2n_3)$
and $O(n_1(r+n_2)n_3\log(n_3)+rn_1n_2n_3)$, respectively.
Note that $r\leq\min\{n_1,n_2\}$.
If we take one of the proximal mappings in Remark \ref{theProMap} for $\mathcal{X}^{k+1}$,
the total cost of ADMM at each iteration is $O(n_1n_2n_3\log(n_3)+rn_1n_2n_3)$.
\end{remark}

\begin{remark}
We propose an ADMM based algorithm to solve model  (\ref{model}),
which can address the general noise model if the proximal mapping of
$-\frac{1}{2\rho}\log p_{\mathcal{X}_\Omega}(\mathcal{Y}_{\Omega})$ can be obtained.
For sparse nonnegative Tucker decomposition, Xu \cite{xu2015alternating} presented an alternating proximal gradient method for solving the proposed model,
where the underlying tensor is corrupted by additive Gaussian noise
and the data-fitting term is differentiable in \cite{xu2015alternating}.
However, this algorithm cannot be applied to our general noise model
since the data-fitting term of our proposed model is not differentiable if the underlying tensor is corrupted by additive Laplace noise.
\end{remark}

%

\section{Numerical Results}\label{NumeriExper}
In this section, some numerical experiments are conducted to
demonstrate the effectiveness of the proposed tensor-based method
for sparse NTF and completion with different noise observations,
including additive Gaussian noise, additive Laplace noise, and Poisson observations.
We will compare the sparse NTF and completion method with the matrix-based method in \cite{soni2016noisy}.

The Karush-Kuhn-Tucker (KKT) conditions of (\ref{ModelOtheFo}) are given by
\begin{equation}\label{KKTCon}
\left\{
\begin{array}{ll}
0\in \partial_\mathcal{X}\left(-\log(p_{\mathcal{X}_{\Omega}}(\mathcal{Y}_{\Omega}))\right)
-\mathcal{T}_1+\mathcal{T}_2, \\
\mathcal{T}_1\diamond\mathcal{B}^T +\mathcal{T}_3=0, \ \mathcal{A}^T\diamond\mathcal{T}_1 +\mathcal{T}_4+\mathcal{T}_5 = 0, \\
0\in \partial{\delta_{\mathfrak{X}'}(\mathcal{Q})}-\mathcal{T}_2, \ 0\in \partial{\delta_{\mathfrak{A}}(\mathcal{M})}-\mathcal{T}_3, \\
0\in\partial(\lambda\|\mathcal{N}\|_0)-\mathcal{T}_4, \ 0\in\partial{\delta_{\mathfrak{B}'}(\mathcal{Z})}-\mathcal{T}_5,\\
\mathcal{X}=\mathcal{A}\diamond \mathcal{B},
\mathcal{Q} = \mathcal{X}, \mathcal{M}=\mathcal{A}, \mathcal{N}=\mathcal{B}, \mathcal{Z}=\mathcal{B},
\end{array}
\right.
\end{equation}
where $\partial f(x)$ denotes the subdifferential of $f$ at $x$.
Based on the KKT conditions in (\ref{KKTCon}),
we adopt the following relative residual to measure the accuracy:
$$
\eta_{max}:=\max\{\eta_1,\eta_2,\eta_3,\eta_4,\eta_5,\eta_6\},
$$
where
\[
\begin{split}
& \eta_1=\frac{\|\mathcal{X}-\textup{Prox}_{(-\log(p_{\mathcal{X}_{\Omega}}
(\mathcal{Y}_{\Omega})))}(\mathcal{T}_1-\mathcal{T}_2+\mathcal{X})\|_F}{1+\|\mathcal{X}\|_F
+\|\mathcal{T}_1\|_F+\|\mathcal{T}_2\|_F}, \
 \eta_2 = \frac{\|\mathcal{Q}-\Pi_{\mathfrak{X}'}(\mathcal{T}_2+\mathcal{Q})\|_F}
 {1+\|\mathcal{T}_2\|_F+\|\mathcal{Q}\|_F}, \\
&\eta_3 = \frac{\|\mathcal{M}-\Pi_{\mathfrak{A}}(\mathcal{T}_3
+\mathcal{M})\|_F}{1+\|\mathcal{T}_3\|_F+\|\mathcal{M}\|_F},  \
\eta_4 = \frac{\|\mathcal{N}-\textup{Prox}_{\lambda\|\cdot\|_0}
(\mathcal{T}_4+\mathcal{N})\|_F}{1+\|\mathcal{T}_4\|_F+\|\mathcal{N}\|_F}, \\
& \eta_5 = \frac{\|\mathcal{Z}-\Pi_{\mathfrak{B}'}(\mathcal{T}_5+\mathcal{Z})\|_F}{1+\|\mathcal{T}_5\|_F+\|\mathcal{Z}\|_F}, \
\eta_6 = \frac{\|\mathcal{X}-\mathcal{A}\diamond \mathcal{B}\|_F}{1+\|\mathcal{X}\|_F + \|\mathcal{A}\|_F + \|\mathcal{B}\|_F}.
\end{split}
\]
Algorithm 1 is terminated if $\eta_{max}<=10^{-4}$ or the number of iterations researches the maximum of $300$.

In order to measure the quality of the recovered tensor,
the relative error (RE) is used to evaluate the performance of different methods,
which is defined as
$$
\textup{RE}=\frac{\|\widetilde{\mathcal{X}}-\mathcal{X}^*\|_F}{\|\mathcal{X}^*\|_F},
$$
where $\widetilde{\mathcal{X}}$ and $\mathcal{X}^*$ are the recovered tensor and the ground-truth tensor, respectively.

\subsection{Synthetic Data}

We generate the nonnegative tensors $\mathcal{A}^*\in\mathbb{R}_+^{n_1\times r\times n_3}$
and $\mathcal{B}^*\in\mathbb{R}_{+}^{r\times n_2\times n_3}$ at random.
$\mathcal{A}^*$ is generated by the MALTAB command $\textup{rand}(n_1,r,n_3)$
and $\mathcal{B}^*$ is a nonnegative sparse tensor generated by the tensor toolbox
command $b\cdot\textup{sptenrand}([r,n_2,n_3],s)$ \cite{TTB_Software}, where $b$ is the magnitude of
$\mathcal{B}^*$ and  $s$ is the sparse ratio.
Then $\mathcal{X}^*=\mathcal{A}^*\diamond \mathcal{B}^*$ and we choose $c=2\|\mathcal{X}^*\|_\infty$.
The size of the testing  third-order tensors is $n_1=n_2=n_3=100$ in the following two  experiments.
The initial values will also influence the performance of ADMM. For the initial values
$\mathcal{A}^0,\mathcal{B}^0, \mathcal{M}^0, \mathcal{N}^0, \mathcal{Z}^0, \mathcal{T}_3^0,  \mathcal{T}_4^0,  \mathcal{T}_5^0$ of ADMM,
we choose them as random tensors with the same size as that of $\mathcal{A}^*$ or $\mathcal{B}^*$.
For the initial values $\mathcal{Q}^0, \mathcal{T}_1^0,  \mathcal{T}_2^0$, we choose them as the observations $\mathcal{Y}_\Omega$ in $\Omega$
and zeros outside $\Omega$.

\begin{figure}[!t]
	\centering
	\subfigure{
		\begin{minipage}[b]{0.99\textwidth}
			\centerline{\scriptsize } \vspace{1.5pt}
			\includegraphics[width=6.3in,height=2.2in]{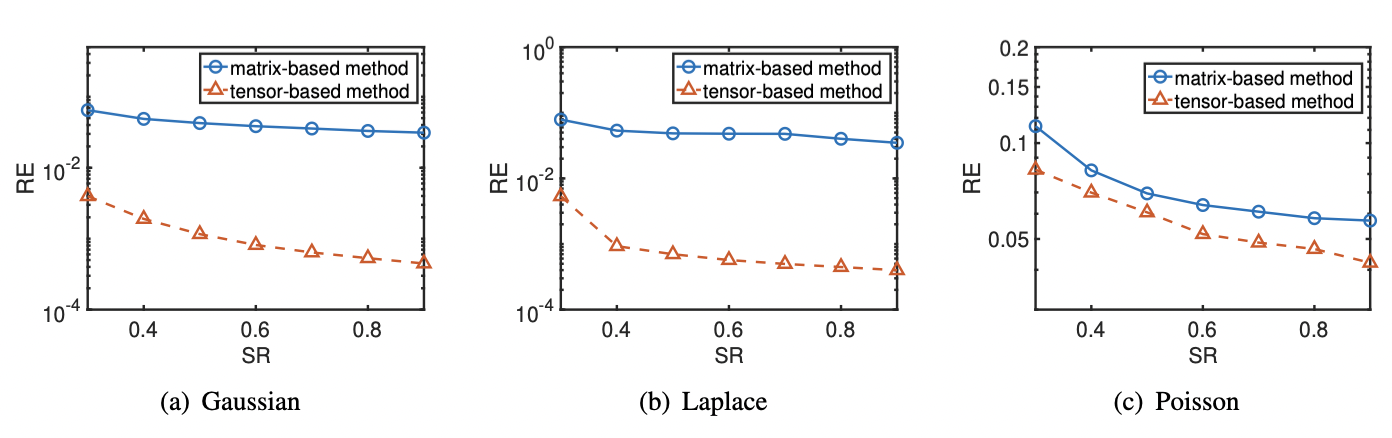}
		\end{minipage}
	}
	\caption{\small RE versus SR of different methods for different noise observations. (a) Gaussian. (b) Laplace. (c) Poisson.}\label{Diffrennoise}
\end{figure}

\begin{figure}[!t]
	\centering
	\subfigure{
		\begin{minipage}[b]{0.99\textwidth}
			\centerline{\scriptsize } \vspace{1.5pt}
			\includegraphics[width=6.3in,height=2.1in]{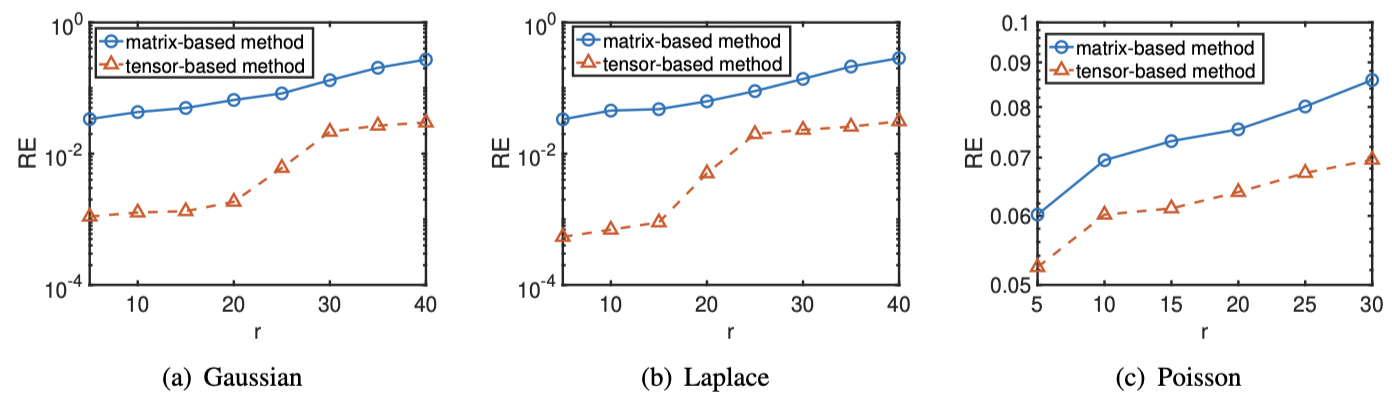}
		\end{minipage}
	}
	\caption{\small RE versus $r$ of different methods for different noise observations.   (a) Gaussian. (b) Laplace. (c) Poisson.}\label{DiffrennoiseR}
\end{figure}

As discussed in Section \ref{ProMod}, we aim to estimate the ground-truth
tensor $\mathcal{X}^*$. Note that  the two factors may not be unique.
Therefore, we only compare the recovered tensor
 $\widetilde{\mathcal{X}}$ with the ground-truth tensor $\mathcal{X}^*$ in the experiments.
 In fact, we just establish the error upper bound between $\widetilde{\mathcal{X}}$ and $\mathcal{X}^*$,
 and do not establish the error bounds of each factor tensor independently in theory.

First we analyze the recovery performance of different methods versus SRs.
In Figure \ref{Diffrennoise}, we display the REs of the recovered tensors with different sampling ratios,
where the sparse ratio $s=0.3$ and the observed entries are corrupted
by additive Gaussian noise, additive Laplace noise, and Poisson noise, respectively.
We set $\sigma=0.1$ and $\tau=0.1$ for additive Gaussian noise and Laplace noise, respectively,
and $r=10$ and $b=2$.
The SRs vary from $0.3$ to $0.9$ with step size $0.1$.
It can be seen from this figure that the REs decrease
when the sampling ratios increase for both matrix- and tensor-based methods.
Moreover, the REs obtained by the tensor-based method are lower than those
obtained by the matrix-based method.
Compared with the matrix-based method, the improvements of the
tensor-based method for additive Gaussian noise and Laplace noise
are much more than those for Poisson observations,
where the main reason is that the constants of the upper error bound in Proposition \ref{uppPoissobs}
are slightly larger than those of the matrix-based method in \cite{soni2016noisy} for Poisson observations.

\begin{figure}[!t]
	\centering
	\subfigure{
		\begin{minipage}[b]{0.99\textwidth}
			\centerline{\scriptsize } \vspace{1.5pt}
			\includegraphics[width=6.3in,height=2in]{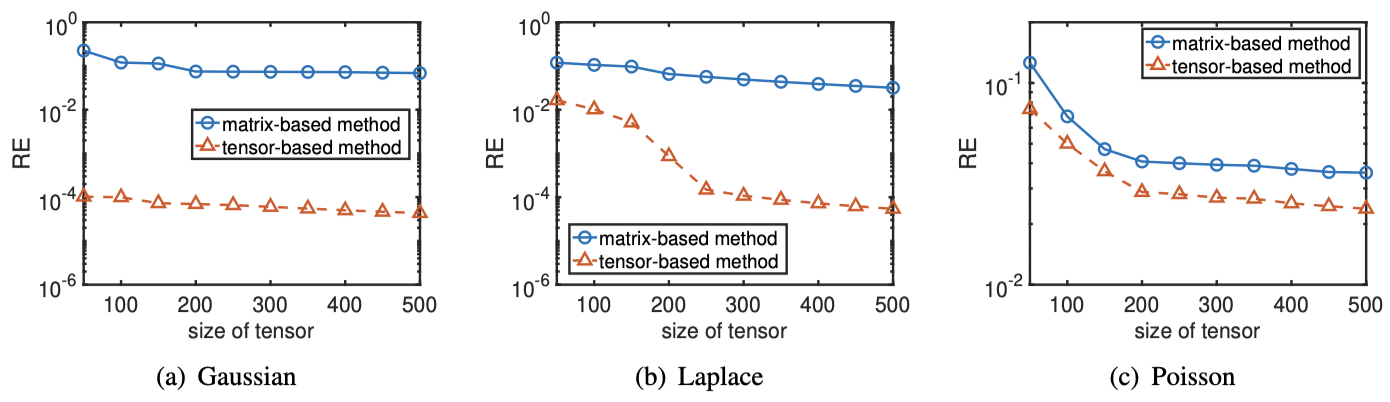}
		\end{minipage}
	}
	\caption{\small RE versus size of tensors of different methods for different noise observations.
  (a) Gaussian. (b) Laplace. (c) Poisson.}\label{DiffrennoiseSize}
\end{figure}

The recovery performance of different methods versus $r$ is discussed in Figure \ref{DiffrennoiseR},
where $\textup{SR} = 0.5$
and $r$ varies from $5$ to $40$ with step size $5$ for additive Gaussian noise and Laplace noise,
and from $5$ to $30$ with step size $5$ for Poisson observations.
Again we set $b=2$, and $\sigma=0.1$ and $\tau=0.1$ for additive  Gaussian noise and Laplace noise, respectively.
It can be seen from this figure that
the REs
obtained by the tensor-based method are lower than those obtained
by the matrix-based method for the three noise models.
Besides, we can observe that the REs increase when $r$ increases for both matrix- and tensor-based methods.
Again the tensor-based method performs better than the matrix-based method for different $r$ and noise observations in terms of REs.
Compared with Poisson observations, the improvements of the tensor-based method are
much more for additive Gaussian noise and Laplace noise.

In Figure \ref{DiffrennoiseSize}, we test different sizes $n:=n_1=n_2=n_3$ of tensors,
and vary $n$ from $50$ to $500$ with step size $50$, where $\textup{SR}=0.5$, the sparse ratio $s=0.3$, $r=10$, and $b=2$.
Here  we set $\sigma=0.1$ for additive Gaussian noise and $\tau=0.1$ for additive Laplace noise, respectively.
It can be observed from this figure that the REs of the tensor-based method are smaller
than those of the matrix-based method for different noise distributions.
The REs of both matrix- and tensor-based methods decrease as $n$ increases.
Furthermore, for different size $n$ of the testing tensors, the improvements of REs of the tensor-based method for
additive Gaussian noise and additive Laplace noise  are much more than those for Poisson observations.

\subsection{Real-World Data}

In this subsection, we test one multispectral image
dataset (length $\times$ width $\times$ spectral) called chart toy\footnote{\footnotesize \url{https://www.cs.columbia.edu/CAVE/databases/multispectral/stuff/}}.
Since this dataset is too large, each image is resized to $128\times 128$
and the resulting tensor is $128\times 128\times 31$  in our experiments.
The underlying image is rescaled to $[0,1]$.

\begin{figure}[!t]
	\centering
	\subfigure{
		\begin{minipage}[b]{0.99\textwidth}
			\centerline{\scriptsize } \vspace{1.5pt}
			\includegraphics[width=6.3in,height=2in]{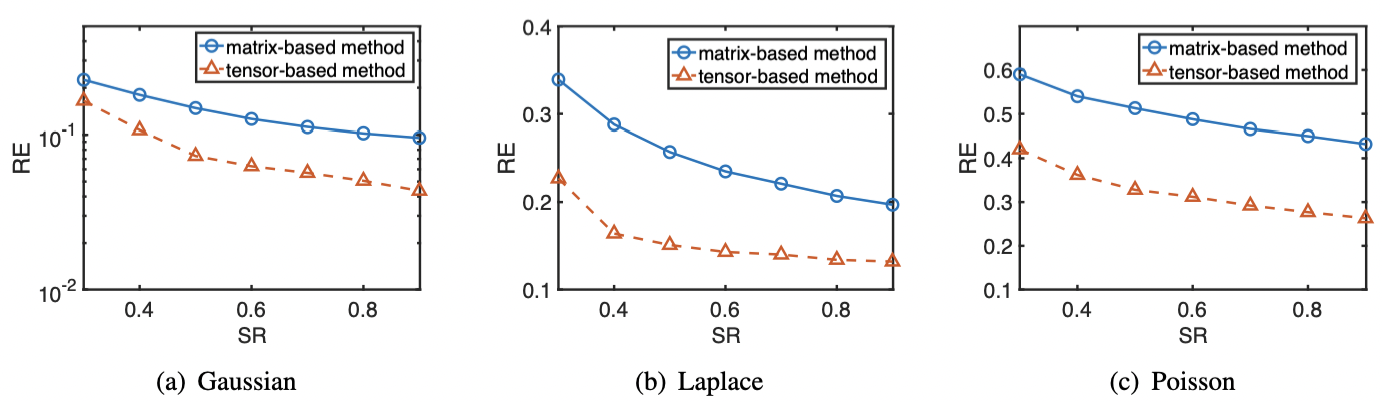}
		\end{minipage}
	}
	\caption{\small RE versus SR of different methods for different noise observations of the chart toy dataset.
  (a) Gaussian. (b) Laplace. (c) Poisson.}\label{RealSR}
\end{figure}

\begin{figure*}[!t]
\centering
\subfigure[Gaussian]{
\begin{minipage}[b]{0.139\textwidth}
\centerline{\scriptsize } \vspace{1.5pt}
\includegraphics[width=1in,height=1in]
{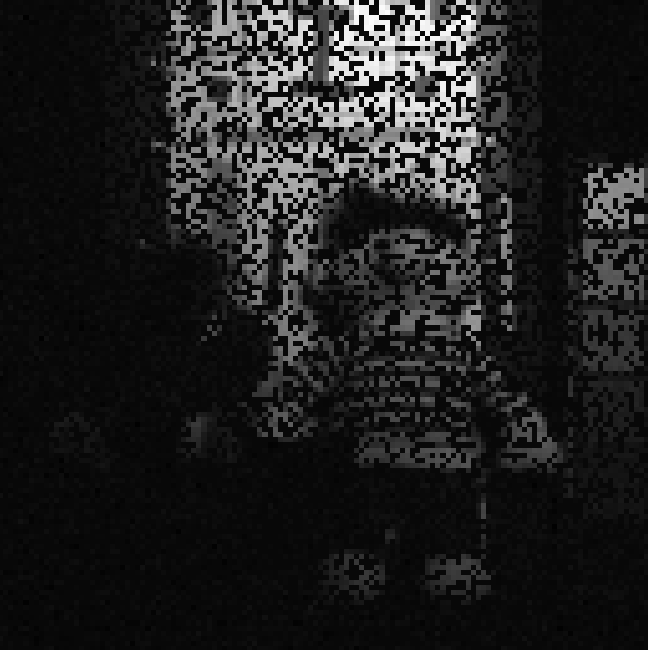}
\centerline{\scriptsize }\\ \vfill \vspace{-22pt}
\includegraphics[width=1in,height=1in]
{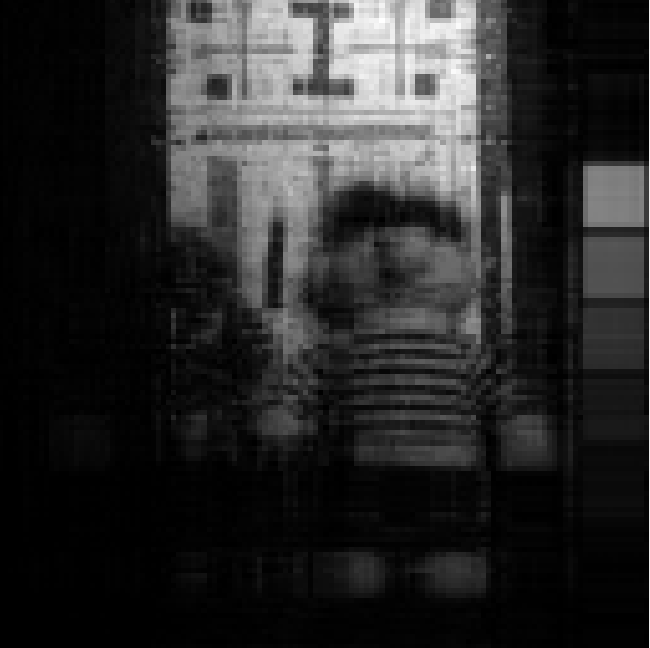}
\centerline{\scriptsize }\\ \vfill \vspace{-22pt}
\includegraphics[width=1in,height=1in]{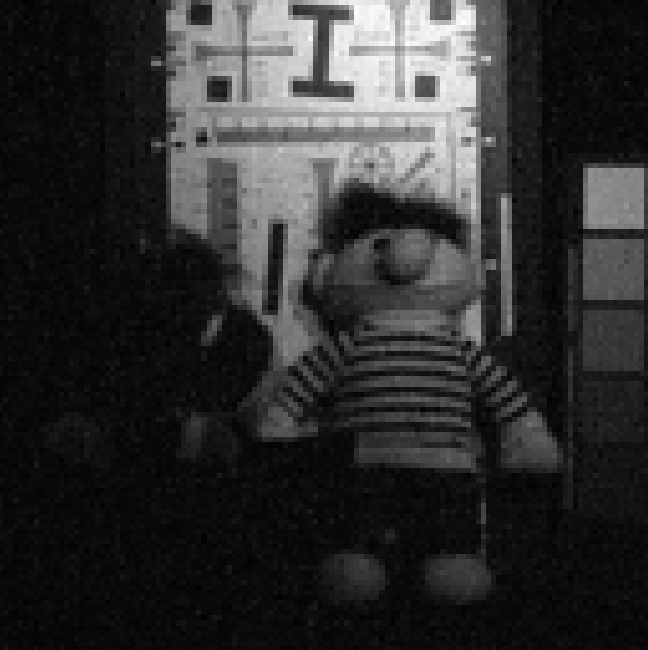}
\end{minipage}
}
\subfigure[Laplace]{
\begin{minipage}[b]{0.139\textwidth}
\centerline{\scriptsize } \vspace{1.5pt}
\includegraphics[width=1in,height=1in]
{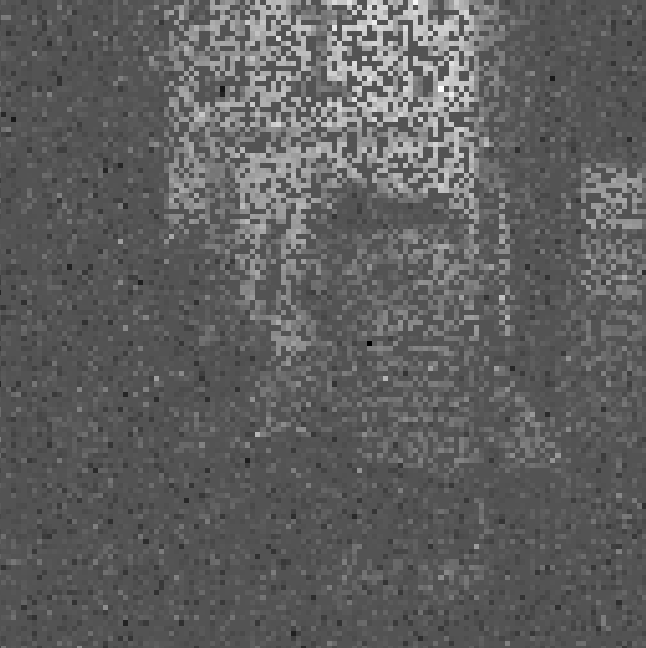}
\centerline{\scriptsize }\\ \vfill \vspace{-22pt}
\includegraphics[width=1in,height=1in]
{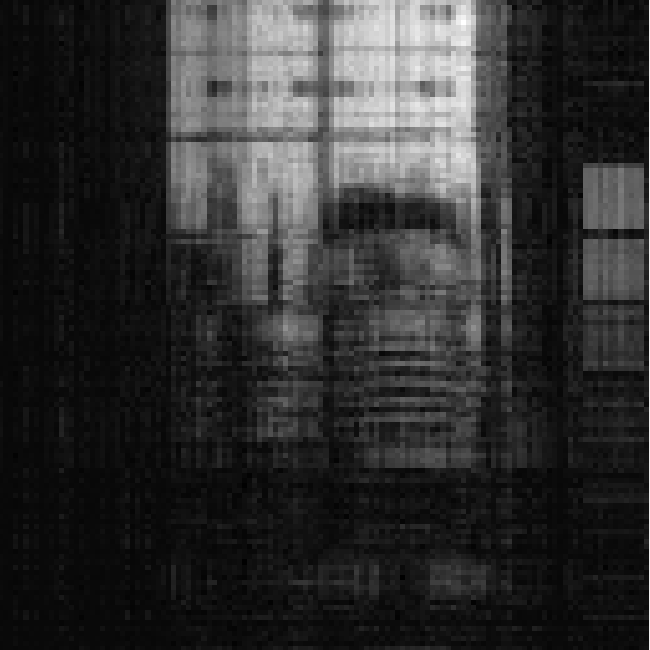}
\centerline{\scriptsize }\\ \vfill \vspace{-22pt}
\includegraphics[width=1in,height=1in]
{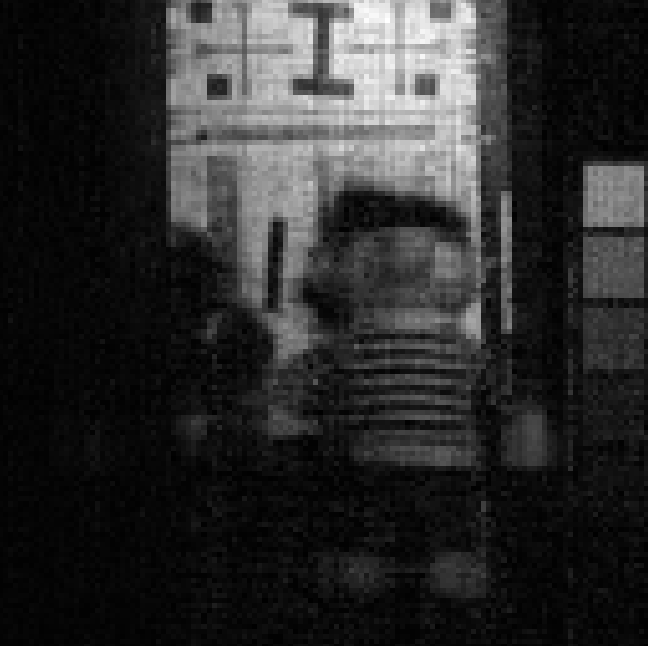}
\end{minipage}
}
\subfigure[Poisson]{
\begin{minipage}[b]{0.139\textwidth}
\centerline{\scriptsize } \vspace{1.5pt}
\includegraphics[width=1in,height=1in]
{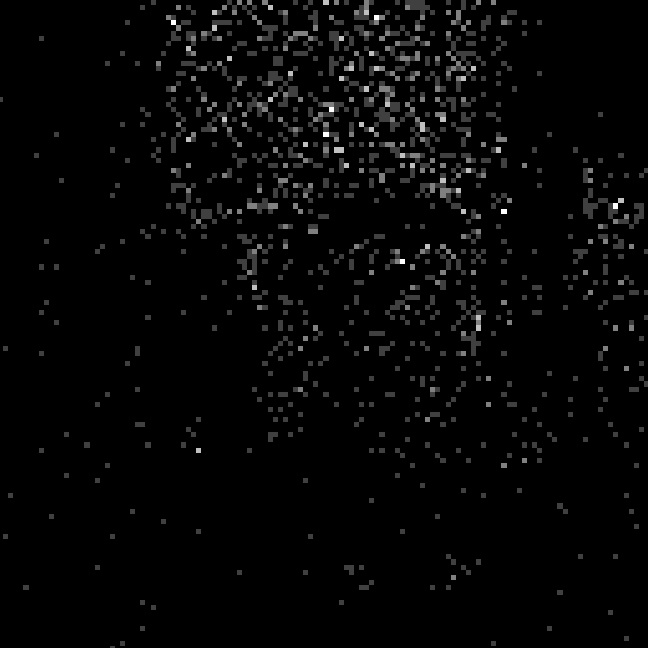}
\centerline{\scriptsize }\\ \vfill \vspace{-22pt}
\includegraphics[width=1in,height=1in]
{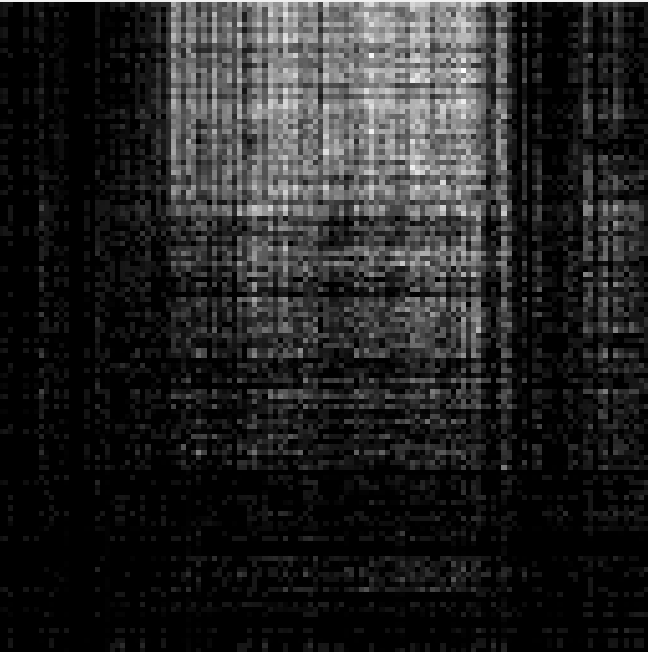}
\centerline{\scriptsize }\\ \vfill \vspace{-22pt}
\includegraphics[width=1in,height=1in]
{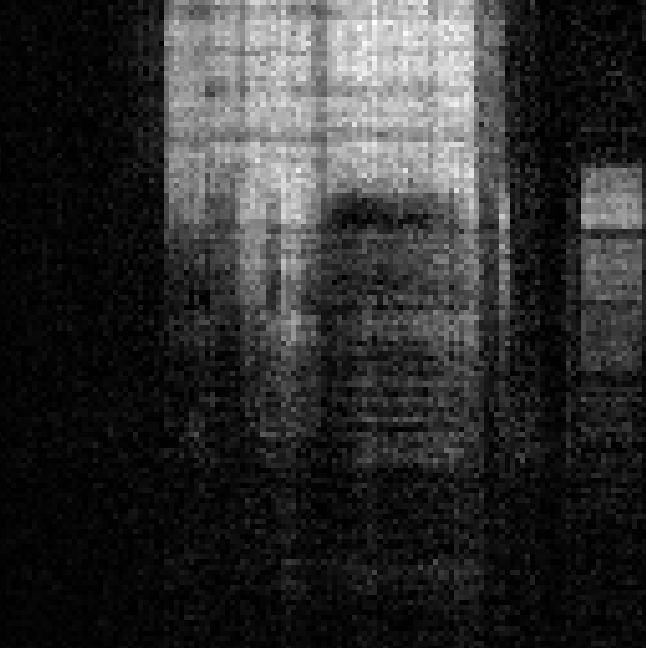}
\end{minipage}
}
\caption{\small  The 25th band of the recovered images
by the matrix- and tensor-based methods for different noise distributions, where $\textup{SR} = 0.6$.
First row: Observed images.
Second row: Recovered images by the matrix-based method.
Third row: Recovered images by the tensor-based method.}\label{RecovImage}
\end{figure*}

Figure \ref{RealSR} shows  the RE versus SR of different methods
for the observations of the chart toy dataset with additive Gaussian noise, additive Laplace noise, and Poisson noise,
where the SR varies from $0.3$ to $0.9$ with step size $0.1$.
The tubal rank of the chart toy dataset is set to $30$ in the experiments.
For additive Gaussian noise and Laplace noise, we set $\sigma=0.1$ and $\tau=0.1$, respectively.
It can be seen from this figure that the REs
obtained by the matrix- and tensor-based methods decrease as the SR increases for different noise distributions.
Moreover, the REs of the tensor-based method are smaller  than those
of the matrix-based method for different sampling ratios and noise distributions.

Figure \ref{RecovImage} shows the visual quality of the 25th band
of the recovered images of different methods and noise distributions, where $\textup{SR}=0.6$.
It can be observed from this figure that the visual quality of
the images recovered by the tensor-based method is better than that recovered by the matrix-based method,
where more details are kept by the tensor-based method.
Furthermore, the recovered images of the two methods for additive Gaussian noise are more clear than those for additive Laplace noise,
which are better than those for Poisson observations in terms of visual quality.

%

\section{Concluding Remarks}\label{Conclu}

In this paper, we have studied the sparse NTF and completion
problem based on tensor-tensor product from partial and noisy observations,
where the observations are corrupted by general noise distributions.
A maximum likelihood estimation of partial observations is derived for the data-fitting term
and the tensor $\ell_0$ norm is adopted to enforce the sparsity of  the sparse factor.
Then an upper error bound is established for a general class of noise models,
and is specialized to widely used noise distributions
including additive Gaussian noise, additive Laplace noise, and Poisson observations.
Moreover, the minimax lower bounds are also established for the previous noise models,
which match the upper error bounds up to logarithmic factors.
An ADMM based algorithm is developed to solve the resulting model.
Preliminary numerical  experiments on both synthetic data and real-world data
are presented to demonstrate the superior performance
of the proposed tensor-based model compared with the matrix-based method \cite{soni2016noisy}.

It would be of great interest to study the upper error bounds of the convexification model
by using the tensor $\ell_1$ norm to replace the tensor $\ell_0$ norm for the sparse factor.
It would also be of great interest to establish  the convergence of ADMM for our proposed model with general noise observations,
which is nonconvex and has multi-block variables.
Moreover, future work may extend the theory of the sparse NTF and completion model with tensor-tensor product to that
with transformed tensor-tensor product under suitable unitary transformations \cite{song2020robust},
which is more effective than tensor-tensor product for robust tensor completion \cite{song2020robust, ng2020patch}
and data compression \cite{zeng2020decompositions}.

\section*{Acknowledgments}
The authors would like to thank
the associate editor and  anonymous referees for their helpful comments
 and constructive suggestions on improving the quality of this paper.

\appendices
\section{Proof of Theorem \ref{maintheo}}\label{ProoA}

We begin by stating the following lemma which will be
useful in the proof of Theorem \ref{maintheo}.

\begin{lemma}\label{leapp}
Let
$\Gamma$ be a countable collection of candidate
reconstructions $\mathcal{X}$ of $\mathcal{X}^*$  and its penalty $\textup{pen}(\mathcal{X})\geq 1$
satisfying $\sum_{\mathcal{X}\in\Gamma}2^{-\textup{pen}(\mathcal{X})}\leq 1$.
For any integer $4\leq m\leq n_1n_2n_3$, let $\Omega\sim  \textup{Bern}(\gamma)$, where $\gamma=\frac{m}{n_1n_2n_3}$.
Moreover, the corresponding observations are obtained by
$p_{\mathcal{X}_{\Omega}^*}(\mathcal{Y}_{\Omega})=\prod_{(i,j,k)\in\Omega}p_{\mathcal{X}_{ijk}^*}(\mathcal{Y}_{ijk})$,
which are assumed to be conditionally independent given $\Omega$.
If
\begin{equation}\label{kappar}
\kappa\geq \max_{\mathcal{X}\in\Gamma}\max_{i,j,k} D(p_{\mathcal{X}_{ijk}^*}(\mathcal{Y}_{ijk})||p_{\mathcal{X}_{ijk}}(\mathcal{Y}_{ijk})),
\end{equation}
then for any
$\xi\geq2\left(1+\frac{2\kappa}{3} \right) \log(2)$,
the following penalized maximum likelihood estimator
\begin{equation}\label{mxial}
\widetilde{\mathcal{X}}^\xi\in\arg\min_{\mathcal{X}\in\Gamma}\left\{-\log p_{\mathcal{X}_\Omega}(\mathcal{Y}_{\Omega})+\xi\cdot\textup{pen}(\mathcal{X})\right\},
\end{equation}
satisfies
$$
\frac{\mathbb{E}_{\Omega,\mathcal{Y}_{\Omega}}\left[-2\log H(p_{\widetilde{\mathcal{X}}^\xi},p_{\mathcal{X}^*})\right]}{n_1n_2n_3}
\leq 3\cdot\min_{\mathcal{X}\in\Gamma}\left\lbrace \frac{D(p_{\mathcal{X}^*}|| p_{\mathcal{X}})}{n_1n_2n_3}+\left( \xi+\frac{4\kappa\log(2)}{3}\right)\frac{\textup{pen}(\mathcal{X})}{m} \right\rbrace +\frac{8\kappa\log(m)}{m},
$$
where the expectation is taken with respect to the joint distribution of $\Omega$ and $\mathcal{Y}_{\Omega}$.
\end{lemma}

The proof of Lemma \ref{leapp} can be derived
easily based on the matrix case \cite[Lemma 8]{soni2016noisy}, see also \cite{li1999estimation}.
At its essence, the three steps of proof in \cite[Lemma 8]{soni2016noisy} are mainly in point-wise manners for
the KL divergence, logarithmic Hellinger affinity, and maximum likelihood estimation.
Therefore, we can extend them to the tensor case easily.
For the sake of brevity, we omit the details here.

Next we give a lemma with respect to the upper bound
of the tensor $\ell_\infty$ norm between a tensor
and its closest surrogate in $\Gamma$.
\begin{lemma}\label{xxappr}
Consider a candidate reconstruction of the form
$\widetilde{\mathcal{X}}^*=\widetilde{\mathcal{A}}^*\diamond \widetilde{\mathcal{B}}^*,$
where each entry of $\widetilde{\mathcal{A}}^*\in\mathfrak{L}$
is the closest discretized surrogates of the entries of $\mathcal{A}^*$,
and each entry of $\widetilde{\mathcal{B}}^*\in\mathfrak{D}$
is the closest discretized surrogates of the nonzero entries of $\mathcal{B}^*$, and zero otherwise.
Then
$$
\|\widetilde{\mathcal{X}}^*-\mathcal{X}^*\|_\infty\leq\frac{3rn_3b}{\vartheta},
$$
where $\vartheta$ is defined as (\ref{denu}).
\end{lemma}
\begin{IEEEproof}
Let $\widetilde{\mathcal{A}}^*=\mathcal{A}^*+\Delta_{\mathcal{A}^*}$
 and $\widetilde{\mathcal{B}}^*=\mathcal{B}^*+\Delta_{\mathcal{B}^*}$.
Then
$$
\widetilde{\mathcal{X}}^*-\mathcal{X}^*
=\widetilde{\mathcal{A}}^*\diamond \widetilde{\mathcal{B}}^*-\mathcal{A}^*\diamond\mathcal{B}^*
=\mathcal{A}^*\diamond\Delta_{\mathcal{B}^*}+\Delta_{\mathcal{A}^*}\diamond\mathcal{B}^*+
\Delta_{\mathcal{A}^*}\diamond\Delta_{\mathcal{B}^*}.
$$
By the definitions of $\widetilde{\mathcal{A}}^*$ and $\widetilde{\mathcal{B}}^*$,
we know that
\begin{equation}\label{DeltaAB}
\|\Delta_{\mathcal{A}^*}\|_\infty\leq \frac{1}{\vartheta-1} \ \
\textup{and} \ \  \|\Delta_{\mathcal{B}^*}\|_\infty\leq \frac{b}{\vartheta-1}.
\end{equation}
Moreover, by the definition of tensor-tensor product of two tensors, we deduce
\begin{equation}\label{ABDSt}
\begin{split}
\mathcal{A}^*\diamond\Delta_{\mathcal{B}^*}& =\textup{Fold}\left(\textup{Circ}\begin{pmatrix} (\mathbf{A^*})^{(1)} \\  (\mathbf{A^*})^{(2)}
\\ \vdots \\  (\mathbf{A^*})^{(n_3)} \end{pmatrix}\cdot \textup{Unfold}(\Delta_{\mathcal{B}^*})\right) \\
&=
\textup{Fold}\begin{pmatrix} (\mathbf{A^*})^{(1)}(\Delta_{\mathcal{B}^*})^{(1)}+(\mathbf{A^*})^{(n_3)}(\Delta_{\mathcal{B}^*})^{(2)}+\cdots +(\mathbf{A^*})^{(2)}(\Delta_{\mathcal{B}^*})^{(n_3)}
\\  (\mathbf{A^*})^{(2)}(\Delta_{\mathcal{B}^*})^{(1)}+ (\mathbf{A^*})^{(1)}(\Delta_{\mathcal{B}^*})^{(2)}+\cdots +  (\mathbf{A^*})^{(3)}(\Delta_{\mathcal{B}^*})^{(n_3)}
\\ \vdots \\  (\mathbf{A^*})^{(n_3)}(\Delta_{\mathcal{B}^*})^{(1)} +(\mathbf{A^*})^{(n_3-1)}(\Delta_{\mathcal{B}^*})^{(2)} +\cdots +(\mathbf{A^*})^{(1)}(\Delta_{\mathcal{B}^*})^{(n_3)} \end{pmatrix}.
\end{split}
\end{equation}
It follows from (\ref{DeltaAB}) and $0\leq \mathcal{A}_{ijk}^*\leq 1$ that
$$
\|\mathcal{A}^*\diamond\Delta_{\mathcal{B}^*}\|_\infty\leq n_3\max_{i,j}\|(\mathbf{A^*})^{(i)} (\Delta_{\mathcal{B}^*})^{(j)}\|_\infty\leq \frac{rn_3b}{\vartheta-1}.
$$
Similarly, we can get that $\|\Delta_{\mathcal{A}^*}\diamond\mathcal{B}^*\|_\infty\leq \frac{rn_3b}{\vartheta-1}$ and $
\|\Delta_{\mathcal{A}^*}\diamond\Delta_{\mathcal{B}^*}\|_\infty\leq \frac{rn_3b}{(\vartheta-1)^2}$.
Therefore, we obtain that
\[
\begin{split}
\|\widetilde{\mathcal{A}}^*\diamond \widetilde{\mathcal{B}}^*-\mathcal{A}^*\diamond\mathcal{B}^*\|_\infty
& \leq \|\mathcal{A}^*\diamond\Delta_{\mathcal{B}^*}\|_\infty+
\|\Delta_{\mathcal{A}^*}\diamond\mathcal{B}^*\|_\infty+
\|\Delta_{\mathcal{A}^*}\diamond\Delta_{\mathcal{B}^*}\|_\infty \\
&\leq \frac{rn_3b}{\vartheta-1}+ \frac{rn_3b}{\vartheta-1}+\frac{rn_3b}{(\vartheta-1)^2}\\
&\leq\frac{3rn_3b}{\vartheta},
\end{split}
\]
 where
the last inequality holds by $\vartheta\geq 8$ in (\ref{denu}).
The proof is completed.
\end{IEEEproof}

\begin{remark}
By the construction of $\widetilde{\mathcal{B}}^*$ in Lemma \ref{xxappr},
we know that $\|\widetilde{\mathcal{B}}^*\|_0=\|\mathcal{B}^*\|_0$,
which will be used to establish the upper bounds in the specifical noise models.
\end{remark}

Now we return to prove Theorem \ref{maintheo}.
First, we need to define the penalty
$\textup{pen}(\mathcal{X})$ on the candidate reconstructions
$\mathcal{X}$ of $\mathcal{X}^*$ in the set $\Gamma_1=\{\mathcal{X}=\mathcal{A}	\diamond \mathcal{B}:
\ \mathcal{A}\in\mathfrak{L}, \ \mathcal{B}\in\mathfrak{D}\}$ such that the summability condition
\begin{equation}\label{penalt}
\sum_{\mathcal{X}\in\Gamma_1}2^{-\textup{pen}(\mathcal{X})}\leq1
\end{equation}
holds, where $\mathfrak{L}$ and $\mathfrak{D}$ are the same as those in $\Gamma$ in (\ref{TauSet}).
Notice that the condition in (\ref{penalt}) is the well-known Kraft-McMillan
inequality for coding entries of $\Gamma$ with an alphabet of size $2$
\cite{Brockway1957Two, kraft1949device}, see also \cite[Section 5]{cover2006elements}.
If we choose the penalties to be code lengths
for some uniquely decodable binary code of the entries $\mathcal{X}\in\Gamma_1$,
then (\ref{penalt}) is satisfied automatically \cite[Section 5]{cover2006elements},
which will provide the constructions of the penalties.

Next we consider the discretized tensor factors $\mathcal{A}\in\mathfrak{L}$
and $ \mathcal{B}\in\mathfrak{D}$.
Fix an ordering of the indices of entries of $\mathcal{A}$
and encode the amplitude of each entry using $\log_2(\vartheta)$ bits.
Let $\widetilde{\vartheta}:=2^{\lceil\log_2(rn_2)\rceil}$.
Similarly, we encode each nonzero entry of $\mathcal{B}$ using $\log_2(\widetilde{\vartheta})$
 bits to denote its location and $\log_2(\vartheta)$ bits for its amplitude.
By this construction, a total of $rn_1n_3\log_2(\vartheta)$ bits
are used to encode $\mathcal{A}$.
Note that $\mathcal{B}$ has $\|\mathcal{B}\|_0$ nonzero entries.
Then a total of $\|\mathcal{B}\|_0(\log_2(\widetilde{\vartheta})+\log_2(\vartheta))$ bits
are used to encode $\mathcal{B}$.
Therefore,  we define the penalties $\textup{pen}(\mathcal{X})$ for all $\mathcal{X}\in\Gamma_1$
as the encoding lengths, i.e.,
$$
\textup{pen}(\mathcal{X})=rn_1n_3\log_2(\vartheta)
+\|\mathcal{B}\|_0(\log_2(\widetilde{\vartheta})+\log_2(\vartheta)).
$$
By the above construction, it is easy to see that such codes are uniquely decodable.
Thus, by Kraft-McMillan inequality \cite{Brockway1957Two, kraft1949device},
we get that $\sum_{\mathcal{X}\in\Gamma_1}2^{-\textup{pen}(\mathcal{X})}\leq1$.
Note that $\Gamma\subseteq \Gamma_1$.
Then  $\sum_{\mathcal{X}\in\Gamma}2^{-\textup{pen}(\mathcal{X})}\leq\sum_{\mathcal{X}\in\Gamma_1}2^{-\textup{pen}(\mathcal{X})}\leq 1$.

Let $\lambda = \xi(\log_2(\vartheta)+\log_2(\widetilde{\vartheta}))$,
where $\xi$ is the regularization parameter in (\ref{mxial}).
Note that
$
\xi\cdot\textup{pen}(\mathcal{X})=\lambda\|\mathcal{B}\|_0+\xi rn_1n_3\log_2(\vartheta).
$
Then the minimizer $\widetilde{\mathcal{X}}^{\lambda}$ in (\ref{model})
is the same as the minimizer $\widetilde{\mathcal{X}}^\xi$ in (\ref{mxial}).
Therefore, by Lemma \ref{leapp},
for any $\xi\geq2\left(1+\frac{2\kappa}{3} \right) \log(2)$,
we get that
\begin{equation}\label{lamxl}
\begin{split}
&\  \frac{\mathbb{E}_{\Omega,\mathcal{Y}_{\Omega}}\left[-2\log H(p_{\widetilde{\mathcal{X}}^\lambda},p_{\mathcal{X}^*})\right]}{n_1n_2n_3}\\
\leq &\ 3\min_{\mathcal{X}\in\Gamma}\left\lbrace \frac{D(p_{\mathcal{X}^*}|| p_{\mathcal{X}})}{n_1n_2n_3}+\left( \xi+\frac{4\kappa\log(2)}{3}\right)\frac{\textup{pen}(\mathcal{X})}{m} \right\rbrace +\frac{8\kappa\log(m)}{m}\\
\leq & \  3\min_{\mathcal{X}\in\Gamma}\left\lbrace \frac{D(p_{\mathcal{X}^*}|| p_{\mathcal{X}})}{n_1n_2n_3}+\left( \xi+\frac{4\kappa\log(2)}{3}\right)\left(\log_2(\vartheta)+\log_2(\widetilde{\vartheta})\right) \frac{rn_1n_3+\|\mathcal{B}\|_0}{m} \right\rbrace +\frac{8\kappa\log(m)}{m} \\
= & \  3\min_{\mathcal{X}\in\Gamma}\left\lbrace \frac{D(p_{\mathcal{X}^*}|| p_{\mathcal{X}})}{n_1n_2n_3}+\left( \lambda+\frac{4\kappa\log(2)}{3}\left(\log_2(\vartheta)+\log_2(\widetilde{\vartheta})\right)\right) \frac{rn_1n_3+\|\mathcal{B}\|_0}{m} \right\rbrace  +\frac{8\kappa\log(m)}{m},
\end{split}
\end{equation}
where the second inequality  holds by the definition of $\textup{pen}(\mathcal{X})$ and the nonnegativity of $\log_2(\vartheta)$ and $\log_2(\widetilde{\vartheta})$.
Note that
\begin{equation}\label{lohvvb}
\log_2(\vartheta)+\log_2(\widetilde{\vartheta})
\leq 2\beta\log_2\left(n_1\vee n_2\right)+2\log_2(rn_2)
\leq\frac{ 2(\beta+2)\log\left(n_1\vee n_2\right)}{\log(2)},
\end{equation}
where the last inequality follows from $rn_2\leq (n_1\vee n_2)^2$.
Hence, for any
$$
\lambda\geq 4(\beta+2)\left( 1+\frac{2\kappa}{3}\right) \log\left(n_1\vee n_2\right),
$$
which is equivalent to $\xi\geq2\left(1+\frac{2\kappa}{3} \right) \log(2)$,
we have
\[
\begin{split}
&\frac{\mathbb{E}_{\Omega,\mathcal{Y}_{\Omega}}[-2\log H(p_{\widetilde{\mathcal{X}}^{\lambda}},p_{\mathcal{X}^*})]}{n_1n_2n_3}\\
\leq & \  3\min_{\mathcal{X}\in\Gamma}\left\lbrace
\frac{D(p_{\mathcal{X}^*}|| p_{\mathcal{X}})}{n_1n_2n_3}+
\left( \lambda+\frac{8\kappa(\beta+2) \log\left(n_1\vee n_2\right)}{3}\right)
\frac{rn_1n_3+\|\mathcal{B}\|_0}{m} \right\rbrace  +\frac{8\kappa\log(m)}{m},
\end{split}
\]
where the inequality follows from (\ref{lamxl}) and (\ref{lohvvb}).
This completes the proof.

\section{Proof of Proposition \ref{Gauuupp}}\label{ProoB}

By Theorem \ref{maintheo}, we only need to establish
the lower bound of $\mathbb{E}_{\Omega,\mathcal{Y}_{\Omega}}
[-2\log H(p_{\widetilde{\mathcal{X}}^{\lambda}},p_{\mathcal{X}^*})]$
and the upper bound of $\min_{\mathcal{X}\in\Gamma}\lbrace
\frac{D(p_{\mathcal{X}^*}|| p_{\mathcal{X}})}{n_1n_2n_3}\rbrace$, respectively.
It follows from \cite[Exercise 15.13]{wainwright2019high}  that
the KL divergence of two Gaussian distributions is  $ D(p_{\mathcal{X}_{ijk}^*}||p_{\mathcal{X}_{ijk}})=(\mathcal{X}_{i,j,k}^*-\mathcal{X}_{i,j,k})^2/(2\sigma^2)$,
which yields
\begin{equation}\label{KLGaussian}
D(p_{\mathcal{X}^*}|| p_{\mathcal{X}})=\frac{\|\mathcal{X}-\mathcal{X}^*\|_F^2}{2\sigma^2}.
\end{equation}
Note that $ D(p_{\mathcal{X}_{ijk}^*}||p_{\mathcal{X}_{ijk}})
\leq c^2/(2\sigma^2)$ for any $\mathcal{X}\in\Gamma$ and $i,j,k$.
Therefore, we can choose $\kappa=c^2/(2\sigma^2)$ based on the assumption in Theorem \ref{maintheo}.
Moreover, by \cite[Appendix C]{carter2002deficiency}, we get that
$$
-2\log(H(p_{\mathcal{X}_{ijk}^*},p_{\widetilde{\mathcal{X}}_{ijk}^\lambda}))
=(\widetilde{\mathcal{X}}_{ijk}^\lambda-\mathcal{X}_{ijk}^*)^2/(4\sigma^2),
$$
which yields that
$
-2\log(H(p_{\mathcal{X}^*},p_{\widetilde{\mathcal{X}}^\lambda}))=\frac{\|\widetilde{\mathcal{X}}^\lambda-\mathcal{X}^*\|_F^2}{4\sigma^2}.
$
As a consequence, by Theorem \ref{maintheo}, we get that
\begin{equation}\label{ErrobG}
\begin{split}
&\frac{\mathbb{E}_{\Omega,\mathcal{Y}_{\Omega}}
[\|\widetilde{\mathcal{X}}^{\lambda}-\mathcal{X}^*\|_F^2]}{n_1n_2n_3}\\
\leq & \  3\min_{\mathcal{X}\in\Gamma}\left\lbrace
\frac{2\|\mathcal{X}-\mathcal{X}^*\|_F^2}{n_1n_2n_3}+
4\sigma^2\left( \lambda+\frac{4c^2(\beta+2) \log(n_1\vee n_2)}{3\sigma^2}\right)
\frac{rn_1n_3+\|\mathcal{B}\|_0}{m} \right\rbrace \\
& +\frac{16c^2\log(m)}{m}.
\end{split}
\end{equation}
Next we need to establish an upper bound of $\min_{\mathcal{X}\in\Gamma}\|\mathcal{X}-\mathcal{X}^*\|_F^2$.
Note that
\begin{equation}\label{vuppb}
\begin{split}
\vartheta&=2^{\lceil\log_2(n_1\vee n_2)^\beta\rceil}\geq 2^{\beta\log_2(n_1\vee n_2)}\\
&\geq2^{\log_2(n_1\vee n_2)}\cdot2^{\log_2(n_1\vee n_2)\frac{\log(3rn_3^{1.5}b/c)}{\log(n_1\vee n_2)}}\\
&=\frac{3(n_1\vee n_2)r{n_3}^{1.5}b}{c},
\end{split}
\end{equation}
where the second inequality holds by (\ref{beta}).
Since $n_1,n_2\geq 2$, we have  $\vartheta\geq\frac{6r{n_3}^{1.5}b}{c}$,
which implies that
$\|\widetilde{\mathcal{X}}^*\|_\infty\leq \frac{3rn_3b}{\vartheta}+\|\mathcal{X}^*\|_\infty\leq c$
by Lemma  \ref{xxappr}, where $\widetilde{\mathcal{X}}^*$ is defined in Lemma \ref{xxappr}.
Therefore, $\widetilde{\mathcal{X}}^*=\widetilde{\mathcal{A}}^*\diamond\widetilde{\mathcal{B}}^*\in\Gamma$.
By Lemma \ref{xxappr}, we have that
\begin{equation}\label{lowxtgm}
\min_{\mathcal{X}\in\Gamma}\left\{\frac{2\|\mathcal{X}-\mathcal{X}^*\|_F^2}{n_1n_2n_3}\right\}
\leq \frac{2\|\widetilde{\mathcal{X}}^*-\mathcal{X}^*\|_F^2}{n_1n_2n_3}
\leq \frac{18(rn_3b)^2}{\vartheta^2}\leq \frac{2c^2}{m},
\end{equation}
where the last inequality follows from the fact $m\leq(n_1\vee n_2)^2n_3$ and (\ref{vuppb}).
Moreover, it follows from the construction of $\widetilde{\mathcal{B}}^*$ in Lemma \ref{xxappr}
that $\|\widetilde{\mathcal{B}}^*\|_0=\|\mathcal{B}^*\|_0$.
As a consequence, combining (\ref{lambda}), (\ref{ErrobG}) with (\ref{lowxtgm}), we obtain that
\[
\begin{split}
&\frac{\mathbb{E}_{\Omega,\mathcal{Y}_{\Omega}}
[\|\widetilde{\mathcal{X}}^{\lambda}-\mathcal{X}^*\|_F^2]}{n_1n_2n_3}\\
\leq & \  \frac{6c^2}{m}+
16(3\sigma^2+2c^2)(\beta+2) \log(n_1\vee n_2)
\left(\frac{rn_1n_3+\|\mathcal{B}^*\|_0}{m}\right)  +\frac{16c^2\log(m)}{m}\\
\leq &  \ \frac{22c^2\log(m)}{m} + 16(3\sigma^2+2c^2)(\beta+2)
\left(\frac{rn_1n_3+\|\mathcal{B}^*\|_0}{m}\right)\log(n_1\vee n_2),
\end{split}
\]
which completes the proof.

\section{Proof of Proposition \ref{AddErp}}\label{ProoC}

A random variable is said to have a Laplace distribution, denoted by  {Laplace}($\mu, b$) with parameters $b>0, \mu$,
if its probability density function is
$
f(x|\mu,b)=\frac{1}{2b}\exp(-\frac{|x-\mu|}{b}).
$
Before deriving the upper bound of observations with additive Laplace noise,
we establish the KL divergence and logarithmic Hellinger affinity between two distributions.
\begin{lemma}\label{KLHeLap}
Let $p(x)\sim\textup{Laplace}(\mu_1, b_1)$ and $q(x)\sim\textup{Laplace}(\mu_2, b_2)$. Then
$$
D(p(x)||q(x))=\log\left(\frac{b_2}{b_1}\right)-1 +\frac{|\mu_2-\mu_1|}{b_2}+\frac{b_1}{b_2}
\exp\left(-\frac{|\mu_2-\mu_1|}{b_1}\right).
$$
Moreover, if $b_1=b_2$, then
$$
-2\log(H(p(x),q(x)))=\frac{|\mu_2-\mu_1|}{b_1}-2\log\left(1+\frac{|\mu_2-\mu_1|}{2b_1}\right).
$$
\end{lemma}
\begin{IEEEproof}
By the definition of the KL divergence of $p(x)$ from $q(x)$, we deduce
\[
\begin{split}
D(p(x)||q(x))&=\mathbb{E}_p\left[\log(p(x))-\log(q(x))\right]\\
&=\log\left(\frac{b_2}{b_1}\right)-\frac{1}{b_1}\mathbb{E}_p[|x-\mu_1|]+\frac{1}{b_2}\mathbb{E}_p[|x-\mu_2|].
\end{split}
\]
Without loss of generality, we assume that $\mu_1<\mu_2$.
By direct calculations, one can get that  $\mathbb{E}_p[|x-\mu_1|]=b_1$ and $\mathbb{E}_p[|x-\mu_2|]=\mu_2-\mu_1+b_1\exp(-\frac{\mu_2-\mu_1}{b_1})$.
Then, we get that
$$
D(p(x)||q(x))=\log\left(\frac{b_2}{b_1}\right)-1+\frac{\mu_2-\mu_1}{b_2}+\frac{b_1\exp(-\frac{\mu_2-\mu_1}{b_1})}{b_2}.
$$
Therefore, by the symmetry, for any $\mu_1,\mu_2$, we have
$$
D(p(x)||q(x))=\log\left(\frac{b_2}{b_1}\right)-1+\frac{|\mu_2-\mu_1|}{b_2}+\frac{b_1\exp\big(-\frac{|\mu_2-\mu_1|}{b_1}\big)}{b_2}.
$$
Moreover, if $b_1=b_2$, the Hellinger affinity is
$$
H(p(x),q(x))=\frac{1}{2b_1}\int_{-\infty}^{+\infty}\exp\left(-\frac{|x-\mu_1|}{2b_1}-\frac{|x-\mu_2|}{2b_1}\right)dx.
$$
With simple manipulations, we obtain
$$
-2\log(H(p(x),q(x)))=\frac{|\mu_2-\mu_1|}{b_1}-2\log\left(1+\frac{|\mu_2-\mu_1|}{2b_1}\right).
$$
The proof is completed.
\end{IEEEproof}

Now we return to prove Proposition \ref{AddErp}.
By Lemma \ref{KLHeLap}, we have that
\begin{equation}\label{KLLappo}
 D(p_{\mathcal{X}_{ijk}^*}||p_{\mathcal{X}_{ijk}})
=\frac{|\mathcal{X}_{ijk}^*-\mathcal{X}_{ijk}|}{\tau}
-\left(1-\exp\left(-\frac{|\mathcal{X}_{ijk}^*-\mathcal{X}_{ijk}|}{\tau}\right)\right)
\leq\frac{1}{2\tau^2}(\mathcal{X}_{ijk}^*-\mathcal{X}_{ijk})^2,
\end{equation}
where the inequality follows from the fact that $e^{-x}\leq 1-x+\frac{x^2}{2}$ for any $x\geq 0$.
Hence, we choose $\kappa=\frac{c^2}{2\tau^2}$ in (\ref{kappar}).
Notice that
\[
\begin{split}
-2\log(H(p_{\mathcal{X}_{ijk}^*},p_{\mathcal{X}_{ijk}}))
&=\frac{|\mathcal{X}_{ijk}^*-\mathcal{X}_{ijk}|}{\tau}
-2\log\left(1+\frac{|\mathcal{X}_{ijk}^*-\mathcal{X}_{ijk}|}{2\tau}\right)\\
&\geq  \frac{(\mathcal{X}_{ijk}^*-\mathcal{X}_{ijk})^2}{(2\tau+c)^2},
\end{split}
\]
where the last inequality follows from the fact that $f(x)=f(0)+f'(0)x+\frac{f''(\xi_0)}{2}x^2$ with $\xi_0\in[0,x]\subseteq[0,c]$ and
$f''(\xi_0)\geq \frac{2}{(2\tau+c)^2}$,
see also  the proof of Corollary 5 in \cite{soni2016noisy}.
Therefore, we have
$$
D(p_{\mathcal{X}^*}||p_{\mathcal{X}})\leq \frac{1}{2\tau^2}\|\mathcal{X}^*-\mathcal{X}\|_F^2
$$
and
\[
\begin{split}
-2\log(H(p_{\mathcal{X}^*},p_{\mathcal{X}}))\geq  \frac{1}{(2\tau+c)^2}\|\mathcal{X}^*-\mathcal{X}\|_F^2.
\end{split}
\]
It follows from Theorem \ref{maintheo} that
\begin{equation}\label{ErLLP}
\begin{split}
&\frac{\mathbb{E}_{\Omega,\mathcal{Y}_{\Omega}}
[\|\widetilde{\mathcal{X}}^{\lambda}-\mathcal{X}^*\|_F^2]}{n_1n_2n_3}\\
\leq & \  3(2\tau+c)^2\cdot\min_{\mathcal{X}\in\Gamma}\left\lbrace
\frac{\|\mathcal{X}^*-\mathcal{X}\|_F^2}{2\tau^2 n_1n_2n_3}+
\left( \lambda+\frac{4c^2(\beta+2) \log(n_1\vee n_2)}{3\tau^2}\right)
\frac{rn_1n_3+\|\mathcal{B}\|_0}{m} \right\rbrace  \\
&\ +\frac{4c^2(2\tau+c)^2\log(m)}{m\tau^2}.
\end{split}
\end{equation}
For the discretitzed surrogate $\widetilde{\mathcal{X}}^*$ of $\mathcal{X}^*$, by Lemma \ref{xxappr}, we get
$$
\min_{\mathcal{X}\in\Gamma}\left\{
\frac{\|\mathcal{X}^*-\mathcal{X}\|_F^2}{2\tau^2 n_1n_2n_3}\right\}
\leq \frac{\|\widetilde{\mathcal{X}}^*-\mathcal{X}^*\|_F^2}{2\tau^2 n_1n_2n_3}
\leq \frac{(3rn_3b)^2}{2\tau^2\vartheta^2}
\leq \frac{c^2}{2\tau^2(n_1\vee n_2)^2n_3}\leq\frac{c^2}{2\tau^2m},
$$
where the third inequality follows from (\ref{vuppb}) and the last inequality
follows from the fact that $m\leq (n_1\vee n_2)^2n_3$.
Note that $\|\widetilde{\mathcal{B}}^*\|_0=\|\mathcal{B}^*\|_0$ by the construction of $\widetilde{\mathcal{X}}^*$ in Lemma \ref{xxappr}.
Combining (\ref{ErLLP}) with (\ref{lambda}), we obtain that
\[
\begin{split}
&\frac{\mathbb{E}_{\Omega,\mathcal{Y}_{\Omega}}
[\|\widetilde{\mathcal{X}}^{\lambda}-\mathcal{X}^*\|_F^2]}{n_1n_2n_3}\\
\leq & \  \frac{3c^2(2\tau+c)^2}{2m\tau^2}
+12\left(1+\frac{2c^2}{3\tau^2}\right)(2\tau+c)^2(\beta+2)\log\left(n_1\vee n_2\right)
\left(\frac{rn_1n_3+\|\mathcal{B}^*\|_0}{m}\right)  \\
&+\frac{4c^2(2\tau+c)^2\log(m)}{m\tau^2}\\
\leq & \ \frac{11c^2(2\tau+c)^2\log(m)}{2m\tau^2}+4\left(3+\frac{2c^2}{\tau^2}\right)(2\tau+c)^2(\beta+2)
\left(\frac{rn_1n_3+\|\mathcal{B}^*\|_0}{m}\right)\log\left(n_1\vee n_2\right).
\end{split}
\]
This completes the proof.

\section{Proof of Proposition \ref{uppPoissobs}}\label{ProoD}

For the KL divergence of Poisson observations,
it follows from \cite[Lemma 8]{cao2016Poisson}  that
\begin{equation}\label{DKLPoi}
D(p_{\mathcal{X}_{ijk}^*}||p_{\mathcal{X}_{ijk}})
\leq \frac{1}{\mathcal{X}_{ijk}}(\mathcal{X}_{ijk}^*-\mathcal{X}_{ijk})^2\leq \frac{1}{\zeta}(\mathcal{X}_{ijk}^*-\mathcal{X}_{ijk})^2.
\end{equation}
Then we can choose $\kappa=\frac{(c-\zeta)^2}{\zeta}$.
Note that
\[
\begin{split}
(\mathcal{X}_{ijk}^*-\mathcal{X}_{ijk})^2
=\left(\left(\sqrt{\mathcal{X}_{ijk}^*}-
\sqrt{\mathcal{X}_{ijk}}\right)\left(\sqrt{\mathcal{X}_{ijk}^*}+\sqrt{\mathcal{X}_{ijk}}\right)\right)^2
\leq 4c(\sqrt{\mathcal{X}_{ijk}^*}-
\sqrt{\mathcal{X}_{ijk}})^2.
\end{split}
\]
Therefore, by \cite[Appendix IV]{raginsky2010compressed}, we have
\begin{equation}\label{PoAil}
\begin{split}
-2\log(H(p_{\mathcal{X}_{ijk}^*},p_{\mathcal{X}_{ijk}}))
&=\left(\sqrt{\mathcal{X}_{ijk}^*}-\sqrt{\mathcal{X}_{ijk}}\right)^2\geq \frac{1}{4c}\left(\mathcal{X}_{ijk}^*-\mathcal{X}_{ijk}\right)^2.
\end{split}
\end{equation}
Therefore, we get
$D(p_{\mathcal{X}^*}||p_{\mathcal{X}})\leq \frac{\|\mathcal{X}^*-\mathcal{X}\|_F^2}{\zeta}$
and
$
-2\log(A(p_{\mathcal{X}^*},p_{\mathcal{X}}))\geq \frac{1}{4c}\|\mathcal{X}^*-\mathcal{X}\|_F^2.
$
For the discreteized surrogate $\widetilde{\mathcal{X}}^*
=\widetilde{\mathcal{A}}^*\diamond\widetilde{\mathcal{B}}^*$ of $\mathcal{X}^*$, by Lemma \ref{xxappr},  we have
\begin{equation}\label{KLPois}
\min_{\mathcal{X}\in\Gamma}\left\{\frac{\|\mathcal{X}-\mathcal{X}^*\|_F^2}{\zeta n_1n_2n_3}\right\}
\leq \frac{\|\widetilde{\mathcal{X}}^*-\mathcal{X}^*\|_F^2}{\zeta n_1n_2n_3}
\leq \frac{9(rn_3b)^2}{\zeta\vartheta^2}\leq \frac{c^2}{\zeta(n_1\vee n_2)^2n_3}\leq \frac{c^2}{\zeta m},
\end{equation}
where the third inequality follows from (\ref{vuppb}).
By the construction of $\widetilde{\mathcal{B}}^*$, we know that  $\|\widetilde{\mathcal{B}}^*\|_0=\|\mathcal{B}^*\|_0$.
Therefore, combining (\ref{PoAil}), (\ref{KLPois}),
and Theorem  \ref{maintheo},  we conclude
\[
\begin{split}
& \ \frac{\mathbb{E}_{\Omega,\mathcal{Y}_{\Omega}}[\|\widetilde{\mathcal{X}}^\lambda-{\mathcal{X}}^*\|_F^2]}{n_1n_2n_3}\\
\leq & \  \frac{12c^3}{\zeta m}+
12c\left( \lambda+\frac{8\kappa(\beta+2) \log\left(n_1\vee n_2\right)}{3}\right)
\frac{rn_1n_3+\|\mathcal{B}^*\|_0}{m} +\frac{32c(c-\zeta)^2\log(m)}{\zeta m}\\
=& \  \frac{4c(3c^2+8(c-\zeta)^2\log(m))}{\zeta m}+
48c\left(1+\frac{4(c-\zeta)^2}{3\zeta}\right)
\frac{(\beta+2) \left(rn_1n_3+\|\mathcal{B}^*\|_0\right)\log\left(n_1\vee n_2\right)}{m} \\
\leq & \ \frac{32c(2c-\zeta)^2\log(m)}{\zeta m}+
48c\left(1+\frac{4(c-\zeta)^2}{3\zeta}\right)
\frac{(\beta+2) \left(rn_1n_3+\|\mathcal{B}^*\|_0\right)\log\left(n_1\vee n_2\right)}{m},
\end{split}
\]
where the equality follows from (\ref{lambda}).
The proof is completed.

\section{Proof of Theorem \ref{lowbounMai}}\label{ProoE}

Let
\begin{equation}\label{GenesubX}
\mathfrak{X}:=\{\mathcal{X}=\mathcal{A}\diamond\mathcal{B}:
\mathcal{A}\in\mathfrak{C},\mathcal{B}\in\mathfrak{B}\},
 \end{equation}
where $\mathfrak{C}\subseteq\mathbb{R}^{n_1\times r\times n_3}$ is  defined as
\begin{equation}\label{CXZG}
\mathfrak{C}:=\left\{\mathcal{A}\in\mathbb{R}^{n_1\times r\times n_3}: \ \mathcal{A}_{ijk}\in\{0,1,a_0\}\right\} \ \ \text{with} \ \  a_0=\min\left\{1, \ \frac{\beta_a\nu}{b\sqrt{\Delta}}\sqrt{\frac{rn_1n_3}{m}}\right\},
\end{equation}
 and $\mathfrak{B}$ is defined as
\begin{equation}\label{GenesubXB}
\mathfrak{B}:=\left\{\mathcal{B}\in\mathbb{R}^{r\times n_2\times n_3}: \
 \mathcal{B}_{ijk}\in\{0,b,b_0\}, \|\mathcal{B}\|_0\leq s\right\}
\ \ \text{with} \  \ b_0=\min\left\{b, \ \frac{\beta_b\nu}{\sqrt{\Delta}}\sqrt{\frac{s}{m}}\right\}.
\end{equation}
Here $\Delta$ is defined as (\ref{deltasp}) and  $\beta_a,\beta_b>0$ are two constants which will be defined later.
From this construction, we get that $\mathfrak{X}\subseteq \mathfrak{U}(r,b,s)$.

Now we define a subset $\mathfrak{X}_{\mathcal{A}}$ such that $\mathfrak{X}_{\mathcal{A}}\subseteq\mathfrak{X}$.
Denote
\begin{equation}\label{SubXAA}
\mathfrak{X}_{\mathcal{A}}:=\left\{\mathcal{X}:=\mathcal{A}\diamond\widetilde{\mathcal{B}}:
 \mathcal{A}\in\widetilde{\mathfrak{C}}, \
\widetilde{\mathcal{B}}=b(\mathcal{I}_r \ \cdots \ \mathcal{I}_r \ \mathbf{0}_\mathcal{B}) \in \mathfrak{B}\right\},
\end{equation}
where $\widetilde{\mathcal{B}}$ is a block tensor with $\lfloor \frac{s\wedge (n_2n_3)}{rn_3}\rfloor$
blocks $\mathcal{I}_r$, $\mathcal{I}_r\in\mathbb{R}^{r\times r\times n_3}$ is the identity tensor,
$\mathbf{0}_{\mathcal{B}}\in\mathbb{R}^{r\times(n_2- \lfloor \frac{s\wedge (n_2n_3)}{rn_3}\rfloor r)\times n_3}$ is the zero tensor with all entries being zero, and
\begin{equation}\label{GeneXACsub}
\widetilde{\mathfrak{C}}:=\left\{\mathcal{A}\in\mathbb{R}^{n_1\times r\times n_3}:\mathcal{A}_{ijk}\in\{0,a_0\},\
1\leq i\leq n_1, 1\leq j\leq r, 1\leq k\leq n_3\right\}.	
\end{equation}
By the definition of identity tensor, we get
$\|\widetilde{\mathcal{B}}\|_0
= r \lfloor \frac{s\wedge (n_2n_3)}{rn_3}\rfloor
\leq r\frac{s\wedge (n_2n_3)}{rn_3}\leq s$.
It follows from  the construction of $\widetilde{\mathcal{B}}=b(\mathcal{I}_r \
\cdots \ \mathcal{I}_r \ \mathbf{0}_{\mathcal{B}})$ that $\widetilde{\mathcal{B}}\in\mathfrak{B}$.
Therefore, $\mathfrak{X}_{\mathcal{A}}\subseteq\mathfrak{X}$.
By the definition of tensor-tensor product, we have that
\[
\begin{split}
\mathcal{A}\diamond \mathcal{I}_r=\textup{Fold}\left(\begin{pmatrix}
\mathbf{A}^{(1)}   &  \mathbf{A}^{(n_3)} & \cdots &  \mathbf{A}^{(2)} \\
 \mathbf{A}^{(2)}  & \mathbf{A}^{(1)}  & \cdots &  \mathbf{A}^{(3)}\\
 \vdots            &  \vdots           &        \ddots & \vdots \\
\mathbf{A}^{(n_3)} &\mathbf{A}^{(n_3-1)}&\cdots &  \mathbf{A}^{(1)} \end{pmatrix}\cdot
\begin{pmatrix} \mathbf{I}_r \\  \mathbf{0}
\\ \vdots \\  \mathbf{0} \end{pmatrix}\right) =\textup{Fold}\begin{pmatrix} \mathbf{A}^{(1)} \\  \mathbf{A}^{(2)}
\\ \vdots \\ \mathbf{A}^{(n_3)} \end{pmatrix} =\mathcal{A},
\end{split}
\]
where $\mathbf{I}_r$ is the $r\times r$ identity matrix.
Hence, for any $\mathcal{X}\in\mathfrak{X}_\mathcal{A}$,
we have that
\begin{equation}\label{XaAI}
\mathcal{X}= b\mathcal{A}\diamond (\mathcal{I}_r \
\cdots \ \mathcal{I}_r \ \mathbf{0}_{\mathcal{B}})=b(\mathcal{A}\ \cdots \ \mathcal{A} \ \mathbf{0}_{\mathcal{X}}),
\end{equation}
where $\mathbf{0}_{\mathcal{X}}\in\mathbb{R}^{n_1\times(n_2-\lfloor
\frac{s\wedge (n_2n_3)}{rn_3}\rfloor r)\times n_3}$ is a zero tensor.
Notice that each entry of $\mathcal{A}$ is $0$ or $a_0$.
Therefore, by the Varshamov-Gilbert bound \cite[Lemma 2.9]{tsybakov2009},
we have that there exists a subset $\mathfrak{X}_\mathcal{A}^0\subseteq \mathfrak{X}_\mathcal{A}$
with $|\mathfrak{X}_\mathcal{A}^0|\geq 2^{rn_1n_3/8}+1$,
such that for any $\mathcal{X}_i,\mathcal{X}_j\in\mathfrak{X}_\mathcal{A}^0$,
\begin{equation}\label{KLADis}
\begin{split}
\|\mathcal{X}_i-\mathcal{X}_j\|_F^2&\geq \frac{r n_1 n_3}{8}
\left\lfloor \frac{s\wedge (n_2n_3)}{rn_3}\right\rfloor a_0^2b^2\\
&\geq \frac{n_1n_2n_3}{16} \min\left\{b^2\Delta, \beta_a^2\nu^2\frac{rn_1n_3}{m}\right\},
\end{split}
\end{equation}
where the last inequality holds
by $\lfloor x\rfloor\geq \frac{x}{2}$ for any $x\geq 1$.
For any $\mathcal{X}\in\mathfrak{X}_{\mathcal{A}}^0$, we have that
\begin{equation}\label{DKLPr}
\begin{split}
D(p_{\mathcal{X}_\Omega}(\mathcal{Y}_\Omega)||p_{\mathbf{0}_\Omega}(\mathcal{Y}_\Omega))
&=\frac{m}{n_1n_2n_3}\sum_{i,j,k}D(p_{\mathcal{X}_{ijk}}(\mathcal{Y}_{ijk})||p_{\mathbf{0}_{ijk}}(\mathcal{Y}_{ijk}))\\
&\leq \frac{m}{n_1n_2n_3}\sum_{i,j,k}\frac{1}{2\nu^2}|\mathcal{X}_{ijk}|^2\\
&\leq \frac{m}{2\nu^2n_1n_2n_3}(rn_1n_3)\left\lfloor \frac{s\wedge (n_2n_3)}{rn_3}\right\rfloor a_0^2b^2\\
& \leq \frac{m}{2\nu^2} \min\left\{\Delta b^2, \beta_a^2\nu^2\frac{rn_1n_3}{m}\right\},
\end{split}
\end{equation}
where the first inequality follows from (\ref{DKLpq}),
the second inequality follows from (\ref{XaAI}) and $|\mathcal{X}_{ijk}|\leq b\|\mathcal{A}\|_\infty$,
and the last inequality follows from (\ref{CXZG}).
Therefore, combining (\ref{CXZG}) with (\ref{DKLPr}), we get that
$$
\sum_{\mathcal{X}\in\mathfrak{X}_\mathcal{A}^0}
D(p_{\mathcal{X}_\Omega}(\mathcal{Y}_\Omega)||p_{\mathbf{0}_\Omega}(\mathcal{Y}_\Omega))
\leq \left(|\mathfrak{X}_\mathcal{A}^0|-1\right)\frac{\beta_a^2rn_1n_3}{2}\leq \left(|\mathfrak{X}_\mathcal{A}^0|-1\right)
\frac{4\beta_a^2\log(|\mathfrak{X}_{\mathcal{A}}^0|-1)}{\log(2)},
$$
where the last inequality holds by $rn_1n_3\leq 8\log_2(|\mathfrak{X}_{\mathcal{A}}^0|-1)$.
Therefore,
by choosing $0<\beta_a\leq \frac{\sqrt{\alpha_1\log(2)}}{2}$ with $0<\alpha_1<\frac{1}{8}$, we have
$$
\frac{1}{|\mathfrak{X}_\mathcal{A}^0|-1}\sum_{\mathcal{X}\in\mathfrak{X}_\mathcal{A}^0}
D(p_{\mathcal{X}_\Omega}(\mathcal{Y}_\Omega)||p_{\mathbf{0}_\Omega}(\mathcal{Y}_\Omega))\leq \alpha_1\log(|\mathfrak{X}_{\mathcal{A}}^0|-1).
$$
Hence, by \cite[Theorem 2.5]{tsybakov2009}, we deduce
\begin{equation}\label{Aminmax}
\begin{split}
&\inf_{\widetilde{\mathcal{X}}}\sup_{\mathcal{X}^*
\in\mathfrak{X}_{\mathcal{A}}}\mathbb{P}\left(\frac{\|\widetilde{\mathcal{X}}-\mathcal{X}^*\|_F^2}{n_1n_2n_3} \geq\frac{1}{32} \min\left\{b^2\Delta, \beta_a^2\nu^2\frac{rn_1n_3}{m}\right\}\right)\\
\geq & \inf_{\widetilde{\mathcal{X}}}\sup_{\mathcal{X}^*
\in\mathfrak{X}_{\mathcal{A}}^0}\mathbb{P}\left(\frac{\|\widetilde{\mathcal{X}}-\mathcal{X}^*\|_F^2}{n_1n_2n_3} \geq\frac{1}{32} \min\left\{b^2\Delta, \beta_a^2\nu^2\frac{rn_1n_3}{m}\right\}\right)\geq \theta_1,
\end{split}
\end{equation}
where
$$
\theta_1=\frac{\sqrt{|\mathfrak{X}_{\mathcal{A}}^0|-1}}
{1+\sqrt{|\mathfrak{X}_{\mathcal{A}}^0|-1}}\left(1-2\alpha_1
-\sqrt{\frac{2\alpha_1}{\log(|\mathfrak{X}_{\mathcal{A}}^0|-1)}}\right)\in(0,1).
$$

Next we consider the sparse factor tensor and construct the packing set, which is included in $\mathfrak{X}$.
Similar to the previous discussion, we construct $\mathfrak{X}_\mathcal{B}$ as follows:
\begin{equation}\label{SubXBB}
\mathfrak{X}_\mathcal{B}:=\left\{\mathcal{X}=\widetilde{\mathcal{A}}\diamond \mathcal{B}:\ \mathcal{B}\in\widetilde{\mathfrak{B}}\right\},	
\end{equation}
where $\widetilde{\mathcal{A}}$ is a block tensor defined as
$$
\widetilde{\mathcal{A}}:=\begin{pmatrix} \mathcal{I}_{r'} & \mathbf{0}
\\\vdots & \vdots\\ \mathcal{I}_{r'} & \mathbf{0} \\ \mathbf{0}
& \mathbf{0} \end{pmatrix}\in\mathbb{R}^{n_1\times r\times n_3}
$$
and $\widetilde{\mathfrak{B}}$ is a set defined as
\begin{equation}\label{GeneXbSubBb}
\widetilde{\mathfrak{B}}:=\left\{\mathcal{B}\in\mathbb{R}^{r\times n_2\times n_3}: \
\mathcal{B}=\begin{pmatrix} \mathcal{B}_{r'} \\ \mathbf{0} \end{pmatrix},
 \mathcal{B}_{r'}\in\mathbb{R}^{r'\times n_2\times n_3},
  ( \mathcal{B}_{r'})_{ijk}\in\{0,b_0\}, \|\mathcal{B}_{r'}\|_0\leq s \right\}.	
\end{equation}
Here  $r'=\lceil\frac{s}{n_2n_3}\rceil$,
$\mathcal{I}_{r'}\in\mathbb{R}^{r'\times r'\times n_3}$ is the identity tensor,
 there are $\lfloor\frac{n_1}{r'}\rfloor$ block tensors $\mathcal{I}_{r'}$ in $\widetilde{\mathcal{A}}$,
 and $\mathbf{0}$ is a zero tensor with all entries being zero whose dimension can be known from the context.
Thus $\mathfrak{X}_{\mathcal{B}}\subseteq \mathfrak{X}$.
Note that
\[
\begin{split}
\mathcal{I}_{r'} \diamond 	\mathcal{B}_{r'} = \textup{Fold}\left(\begin{pmatrix}
\mathbf{I}_{r'}   &  \mathbf{0} & \cdots &  \mathbf{0} \\
 \mathbf{0}  &\mathbf{I}_{r'}  & \cdots &  \mathbf{0}\\
 \vdots            &  \vdots           &        \ddots & \vdots \\
\mathbf{0} &\mathbf{0} &\cdots &  \mathbf{I}_{r'}  \end{pmatrix}\cdot
\begin{pmatrix} \mathbf{B}_{r'}^{(1)} \\  \mathbf{B}_{r'}^{(2)}
\\ \vdots \\  \mathbf{B}_{r'}^{(n_3)} \end{pmatrix}\right)=
\textup{Fold}
\begin{pmatrix} \mathbf{B}_{r'}^{(1)} \\  \mathbf{B}_{r'}^{(2)}
\\ \vdots \\  \mathbf{B}_{r'}^{(n_3)} \end{pmatrix}=\mathcal{B}_{r'}.
\end{split}
\]
For any $\mathcal{X}\in\mathfrak{X}_{\mathcal{B}}$, we have
$$
\mathcal{X}=\begin{pmatrix} \mathcal{I}_{r'} & \mathbf{0}
\\\vdots & \vdots\\ \mathcal{I}_{r'} & \mathbf{0} \\ \mathbf{0}
& \mathbf{0} \end{pmatrix} \diamond \begin{pmatrix} \mathcal{B}_{r'} \\ \mathbf{0} \end{pmatrix}=\begin{pmatrix} \mathcal{B}_{r'} \\ \vdots \\  \mathcal{B}_{r'} \\ \mathbf{0} \end{pmatrix},
$$
where $ ( \mathcal{B}_{r'})_{ijk}\in\{0,b_0\}$, $\|\mathcal{B}_{r'}\|_0\leq s$,
and there are $\lfloor\frac{n_1}{r'}\rfloor$ blocks $ \mathcal{B}_{r'}$ in $\mathcal{X}$.
By the Varshamov-Gilbert bound \cite[Lemma 2.9]{tsybakov2009},
there is a subset $\mathfrak{X}_{\mathcal{B}}^0\subseteq \mathfrak{X}_{\mathcal{B}}$ such that
for any $\mathcal{X}_i,\mathcal{X}_j\in\mathfrak{X}_{\mathcal{B}}^0$,
\begin{equation}\label{XBKL}
|\mathfrak{X}_{\mathcal{B}}^0|\geq 2^{r'n_2n_3/8}+1 \geq 2^{s/8}+1
\end{equation}
 and
\[
\begin{split}
\|\mathcal{X}_i-\mathcal{X}_j\|_F^2&
\geq  \frac{r'n_2n_3}{8}\left\lfloor\frac{n_1}{r'}\right\rfloor b_0^2
\geq \frac{s}{8}\left\lfloor\frac{n_1}{r'}\right\rfloor b_0^2 \\
& \geq \frac{sn_1}{16r'}b_0^2=\frac{n_1n_2n_3}{16}\frac{s}{n_2n_3\lceil\frac{s}{n_2n_3}\rceil}b_0^2\\
&\geq \frac{n_1n_2n_3}{16}\min\left\{\frac{1}{2},\frac{s}{n_2n_3}\right\}b_0^2
\geq \frac{n_1n_2n_3}{32}\Delta\cdot\min\left\{b^2,\frac{\beta_b^2\nu^2s}{\Delta m}\right\}\\
&=\frac{n_1n_2n_3}{32}\min\left\{\Delta b^2,\frac{\beta_b^2\nu^2s}{m}\right\},
\end{split}
\]
where the third inequality follows from $\lfloor x\rfloor\geq \frac{x}{2}$ for any $x\geq 1$ and
the fourth inequality follows from the fact that $\frac{x}{\lceil x\rceil}\geq \min\{\frac{1}{2},x\}$ for any $x>0$.

For any $\mathcal{X}\in\mathfrak{X}_{\mathcal{B}}^0$,
the KL divergence of the observations with parameter $\mathcal{X}_\Omega$ from the observations with parameter $\mathbf{0}_\Omega$ is given by
\[
\begin{split}
D(p_{\mathcal{X}_\Omega}(\mathcal{Y}_{\Omega})||p_{\mathbf{0}_\Omega}(\mathcal{Y}_{\Omega}))
& =\frac{m}{n_1n_2n_3}\sum_{i,j,k}D(p_{\mathcal{X}_{ijk}}(\mathcal{Y}_{ijk})||p_{\mathbf{0}_{ijk}}(\mathcal{Y}_{ijk}))
\leq \frac{m}{2 \nu^2n_1n_2n_3}\sum_{i,j,k}|\mathcal{X}_{ijk}|^2\\
&\leq \frac{m}{2 \nu^2n_1n_2n_3}n_1(s\wedge (n_2n_3))b_0^2=\frac{m}{2\nu^2}\min\left\{\Delta b^2, \frac{\beta_b^2\nu^2s}{m}\right\}\\
&\leq \frac{\beta_b^2s}{2}\leq 4\beta_b^2\frac{\log(|\mathfrak{X}_\mathcal{B}^0|-1)}{\log(2)},
\end{split}
\]
where the second inequality follows from the fact that the nonzero entries of $\mathcal{X}$ is not larger than
$s\lfloor\frac{n_1}{r'}\rfloor \leq n_1(s\wedge (n_2n_3))$, and the last inequality holds by
$s\leq 8\log_{2}(|\mathfrak{X}_\mathcal{B}^0|-1)$.
By choosing $0<\beta_b\leq \frac{\sqrt{\alpha_2\log(2)}}{2}$ with $0<\alpha_2<\frac{1}{8}$, we obtain that
$$
\frac{1}{|\mathfrak{X}_\mathcal{B}^0|-1}
\sum_{\mathcal{X}\in\mathfrak{X}_\mathcal{B}^0} D(p_{\mathcal{X}_\Omega}(\mathcal{Y}_{\Omega})||p_{\mathbf{0}_\Omega}(\mathcal{Y}_{\Omega}))
\leq\alpha_2\log(|\mathfrak{X}_\mathcal{B}^0|-1).
$$
Therefore, by \cite[Theorem 2.5]{tsybakov2009}, we have that
\begin{equation}\label{Bminmax}
\begin{split}
&\inf_{\widetilde{\mathcal{X}}}\sup_{\mathcal{X}^*
\in\mathfrak{X}_{\mathcal{B}}}\mathbb{P}\left(\frac{\|\widetilde{\mathcal{X}}-\mathcal{X}^*\|_F^2}{n_1n_2n_3} \geq\frac{1}{64} \min\left\{\Delta b^2,\frac{\beta_b^2\nu^2s}{m}\right\}\right)\\
\geq & \inf_{\widetilde{\mathcal{X}}}\sup_{\mathcal{X}^*
\in\mathfrak{X}_{\mathcal{B}}^0}\mathbb{P}\left(\frac{\|\widetilde{\mathcal{X}}-\mathcal{X}^*\|_F^2}{n_1n_2n_3} \geq\frac{1}{64} \min\left\{\Delta b^2,\frac{\beta_b^2\nu^2s}{m}\right\}\right)\geq \theta_2,
\end{split}
\end{equation}
where
$$
\theta_2=\frac{\sqrt{|\mathfrak{X}_\mathcal{B}^0|-1}}
{1+\sqrt{|\mathfrak{X}_\mathcal{B}^0|-1}}\left(1-2\alpha_2
-\sqrt{\frac{2\alpha_2}{\log(|\mathfrak{X}_\mathcal{B}^0|-1)}}\right)\in(0,1).
$$
Let $\beta_c=\min\{\beta_a,\beta_b\}$ and $\theta_c=\min\{\theta_1,\theta_2\}.$
Combining (\ref{Aminmax}) and (\ref{Bminmax}), we deduce
$$
\inf_{\widetilde{\mathcal{X}}}\sup_{\mathcal{X}^*
\in\mathfrak{U}(r,b,k)}\mathbb{P}\left(\frac{\|\widetilde{\mathcal{X}}-\mathcal{X}^*\|_F^2}{n_1n_2n_3} \geq\frac{3}{128} \min\left\{\Delta b^2,\beta_c^2\nu^2\left(\frac{s+rn_1n_3}{m}\right)\right\}\right)\geq \theta_c.
$$
By  Markov's inequality, we conclude
$$
\inf_{\widetilde{\mathcal{X}}}\sup_{\mathcal{X}^*
\in\mathfrak{U}(r,b,k)}\frac{\mathbb{E}_{\Omega,\mathcal{Y}_\Omega}[\|\widetilde{\mathcal{X}}-\mathcal{X}^*\|_F^2]}{n_1n_2n_3}\geq
\frac{3\theta_c}{128}\cdot \min\left\{\Delta b^2,\beta_c^2\nu^2\left(\frac{s+rn_1n_3}{m}\right)\right\}.
$$
This completes the proof.

\section{Proof of Proposition \ref{ProupbG}}\label{AppdeF}

By (\ref{KLGaussian}), we choose $\nu=\sigma$.
It follows from Theorem \ref{lowbounMai} that we can get the desired result.

\section{Proof of Proposition \ref{lapUpb}}\label{AppdeG}

By (\ref{KLLappo}), we can choose $\nu=\tau$.
Then the conclusion can be obtained  easily by Theorem  \ref{lowbounMai}.

\section{Proof of Proposition \ref{Poisslow}}\label{ProoH}

Let \begin{equation}\label{PoscsubX1}
 \mathfrak{X}_1:=\{\mathcal{X}=\mathcal{A}\diamond\mathcal{B}:
\mathcal{A}\in\mathfrak{C}_1,\mathcal{B}\in\mathfrak{B}_1\},
 \end{equation}
where $\mathfrak{C}_1\subseteq\mathbb{R}^{n_1\times r\times n_3}$ is  defined as
\begin{equation}\label{CXZ}
\mathfrak{C}_1:=\left\{\mathcal{A}\in\mathbb{R}^{n_1\times r\times n_3}:\
\mathcal{A}_{ijk}\in\{0,1,\varsigma,a_0\}\right\} \  \text{with} \  a_0=\min\left\{1-\varsigma, \ \frac{\beta_a\sqrt{\zeta}}{b}\sqrt{\frac{rn_1n_3}{m}}\right\},
\end{equation}
and $\mathfrak{B}$ is defined as
\begin{equation}\label{PoscsubX1B1}
\mathfrak{B}_1:=\left\{\mathcal{B}\in\mathbb{R}^{r\times n_2\times n_3}:
 \mathcal{B}_{ijk}\in\{0,\zeta, b,b_0\}, \|\mathcal{B}\|_0\leq s\right\}
\ \text{with}  \ b_0=\min\left\{b, \ \beta_b\sqrt{\frac{\zeta}{\Delta_1}}\sqrt{\frac{s-n_2n_3}{m}}\right\}.	
\end{equation}

We discuss the two factors separately.

{\bf Case I.} Let
\begin{equation}\label{PoissXA1A}
\widetilde{\mathfrak{X}}_\mathcal{A}:=\left\{\mathcal{X}
:=(\mathcal{A}+\mathcal{A}_\varsigma)\diamond\mathcal{B}: \ \mathcal{A}\in\widetilde{\mathfrak{C}}_1, \
\mathcal{B}=b(\mathcal{I}_r \ \cdots  \ \mathcal{I}_r \ \mathcal{B}_\mathcal{I} )\in\mathfrak{B}_1\right\},	
\end{equation}
where $\mathcal{I}_r\in\mathbb{R}^{r\times r\times n_3}$ is the identity tensor,
there are $\lfloor\frac{n_2}{r}\rfloor$ blocks $\mathcal{I}_r$
in $\mathcal{B}$, $\mathcal{B}_\mathcal{I}=\begin{pmatrix} \mathcal{I}_{\mathcal{B}} \\ \mathbf{0} \end{pmatrix}$,
$\mathcal{I}_{\mathcal{B}}\in\mathbb{R}^{(n_2-r\lfloor\frac{n_2}{r}\rfloor)\times (n_2-r\lfloor\frac{n_2}{r}\rfloor)\times n_3}$ is the identity tensor,
$\mathcal{A}_\varsigma\in \mathbb{R}^{n_1\times r\times n_3}$
with $(\mathcal{A}_\varsigma)_{ijk}=\varsigma$, and
\begin{equation}\label{PoisXAsubC1}
\widetilde{\mathfrak{C}}_1:=\left\{\mathcal{A}\in\mathbb{R}^{n_1\times r\times n_3}: \
\mathcal{A}_{ijk}\in\{0,a_0\}\right\}\subseteq \mathfrak{C}_1.	
\end{equation}
From the construction of $\mathcal{B}$, we know that $\|\mathcal{B}\|_0=n_2< s$.
For any $\mathcal{X}\in\widetilde{\mathfrak{X}}_\mathcal{A}$, we obtain that
\begin{equation}\label{XDCL}
\mathcal{X}=(\mathcal{A}+\mathcal{A}_\varsigma)\diamond\mathcal{B}=\zeta\mathbb{I}\diamond(\mathcal{I}_r \ \cdots  \ \mathcal{I}_r \ \mathcal{B}_\mathcal{I} )+\mathcal{A}\diamond\mathcal{B},
\end{equation}
where $\mathbb{I}\in\mathbb{R}^{n_1\times r\times n_3}$
denotes a tensor with all entries being $1$.
By the definition of tensor-tensor product, we have that
\[
\begin{split}
\mathbb{I}\diamond(\mathcal{I}_r \ \cdots  \ \mathcal{I}_r \ \mathcal{B}_\mathcal{I})& =
\textup{Fold}\left(\begin{pmatrix}
\mathbf{E}_{n_1r}   &  \mathbf{E}_{n_1r} & \cdots &  \mathbf{E}_{n_1r} \\
 \mathbf{E}_{n_1r}   &\mathbf{E}_{n_1r}  & \cdots &  \mathbf{E}_{n_1r} \\
 \vdots            &  \vdots           &        \ddots & \vdots \\
\mathbf{E}_{n_1r}  &\mathbf{E}_{n_1r} &\cdots &  \mathbf{E}_{n_1r}  \end{pmatrix}\cdot
\begin{pmatrix}
\mathbf{I}_{r}   &  \mathbf{I}_r & \cdots &   \mathbf{I}_r & \mathbf{I}_{B0} \\
 \mathbf{0}  &\mathbf{0}  & \cdots &\mathbf{0}  &  \mathbf{0}\\
 \vdots            &  \vdots           &        \ddots & \vdots & \vdots \\
\mathbf{0} &\mathbf{0} &\cdots &  \mathbf{0}  &\mathbf{0}  \end{pmatrix}\right)\\
& =
\textup{Fold}
\begin{pmatrix} \mathbf{E}_{n_1n_2} \\  \mathbf{E}_{n_1n_2}
\\ \vdots \\ \mathbf{E}_{n_1n_2} \end{pmatrix}=\mathbb{I}_{n_1n_2},
\end{split}
\]
where $\mathbf{E}_{n_1r}\in \mathbb{R}^{n_1\times r}$ is an $n_1\times r$ matrix with all entries being $1$, $\mathbf{I}_{r}$
is the $r\times r$ identity matrix,
$$
\mathbf{I}_{B0}=\begin{pmatrix} \mathbf{I}_{B} \\ \mathbf{0} \end{pmatrix}\in\mathbb{R}^{r\times (n_2-r\lfloor\frac{n_2}{r}\rfloor)}
$$
with $\mathbf{I}_{B}\in\mathbb{R}^{(n_2-r\lfloor\frac{n_2}{r}\rfloor)\times (n_2-r\lfloor\frac{n_2}{r}\rfloor)}$
being the identity matrix,
and $\mathbb{I}_{n_1n_2}\in\mathbb{R}^{n_1\times n_2 \times n_3}$ is a tensor with all entries being $1$.
Therefore, we have $\widetilde{\mathfrak{X}}_\mathcal{A}\subseteq\widetilde{\mathfrak{U}}(r,b,s,\zeta)$.
By applying the Varshamov-Gilbert
bound \cite[Lemma 2.9]{tsybakov2009} to the last term of (\ref{XDCL}), there is a subset
$\widetilde{\mathfrak{X}}_\mathcal{A}^0\subseteq\widetilde{\mathfrak{X}}_\mathcal{A}$
such that for any $\mathcal{X}_1,\mathcal{X}_2\in\widetilde{\mathfrak{X}}_\mathcal{A}^0$,
$$
\|\mathcal{X}_1-\mathcal{X}_2\|_F^2\geq \frac{rn_1n_3}{8}\left\lfloor\frac{n_2}{r}\right\rfloor a_0^2b^2\geq
\frac{n_1n_2n_3}{16}\min\left\{(1-\varsigma)^2b^2, \ \frac{\beta_a^2\zeta r n_1n_3}{m}\right\}
$$
and $|\widetilde{\mathfrak{X}}_\mathcal{A}^0|\geq 2^{\frac{rn_1n_3}{8}}+1$.
Let $\mathcal{X}_0=\zeta\mathbb{I}\diamond(\mathcal{I}_r \ \cdots  \ \mathcal{I}_r \ \mathcal{B}_\mathcal{I} )$.
For any $\mathcal{X}\in\widetilde{\mathfrak{X}}_\mathcal{A}^0$, the KL divergence of $p_{\mathcal{X}_\Omega}(\mathcal{Y}_{\Omega})$ from $p_{(\mathcal{X}_0)_\Omega}(\mathcal{Y}_{\Omega})$ is given by
\[
\begin{split}
D(p_{\mathcal{X}_\Omega}(\mathcal{Y}_{\Omega})||p_{(\mathcal{X}_0)_\Omega}(\mathcal{Y}_{\Omega}))
&=\frac{m}{n_1n_2n_3}\sum_{i,j,k}D(p_{\mathcal{X}_{ijk}}(\mathcal{Y}_{ijk})||p_{(\mathcal{X}_0)_{ijk}}(\mathcal{Y}_{ijk})) \\
&\leq \frac{m}{n_1n_2n_3}\sum_{i,j,k}\frac{(\mathcal{X}_{ijk}-\zeta)^2}{\zeta}\\
&\leq \frac{m(a_0b)^2}{\zeta}\leq \beta_a^2rn_1n_3,
\end{split}
\]
where the first inequality follows from (\ref{DKLPoi}),
the second inequality follows from (\ref{XDCL}),
and the last inequality follows from (\ref{CXZ}).
Note that $rn_1n_3\leq \frac{8\log(|\widetilde{\mathfrak{X}}_\mathcal{A}^0|-1)}{\log(2)}$.
Then, by choosing $0<\beta_a
\leq \frac{\sqrt{\widetilde{\alpha}_1\log(2)}}{2\sqrt{2}}$ with $0<\widetilde{\alpha}_1<\frac{1}{8}$,
 we get that
$$
\frac{1}{|\widetilde{\mathfrak{X}}_\mathcal{A}^0|-1}\sum_{\mathcal{X}\in\widetilde{\mathfrak{X}}_\mathcal{A}^0}
D(p_{\mathcal{X}_\Omega}(\mathcal{Y}_\Omega)||p_{(\mathcal{X}_0)_\Omega}(\mathcal{Y}_\Omega))\leq \widetilde{\alpha}_1\log(|\widetilde{\mathfrak{X}}_{\mathcal{A}}^0|-1).
$$
Therefore, by \cite[Theorem 2.5]{tsybakov2009}, we have that
\begin{equation}\label{PXKL}
\begin{split}
&\inf_{\widetilde{\mathcal{X}}}\sup_{\mathcal{X}^*
\in\widetilde{\mathfrak{X}}_{\mathcal{A}}}\mathbb{P}\left(\frac{\|\widetilde{\mathcal{X}}-\mathcal{X}^*\|_F^2}{n_1n_2n_3} \geq
\frac{1}{32}\min\left\{(1-\varsigma)^2b^2, \ \frac{\beta_a^2\zeta r n_1n_3}{m}\right\}\right)\\
\geq & \inf_{\widetilde{\mathcal{X}}}\sup_{\mathcal{X}^*
\in\widetilde{\mathfrak{X}}_{\mathcal{A}}^0}\mathbb{P}\left(\frac{\|\widetilde{\mathcal{X}}-\mathcal{X}^*\|_F^2}{n_1n_2n_3} \geq\frac{1}{32} \min\left\{(1-\varsigma)^2b^2, \ \frac{\beta_a^2\zeta r n_1n_3}{m}\right\}\right)\geq \widetilde{\theta}_1,
\end{split}
\end{equation}
where
$$
\widetilde{\theta}_1=\frac{\sqrt{|\widetilde{\mathfrak{X}}_{\mathcal{A}}^0|-1}}
{1+\sqrt{|\widetilde{\mathfrak{X}}_{\mathcal{A}}^0|-1}}\left(1-2\widetilde{\alpha}_1
-\sqrt{\frac{2\widetilde{\alpha}_1}{\log(|\widetilde{\mathfrak{X}}_{\mathcal{A}}^0|-1)}}\right)\in(0,1).
$$

{\bf Case II.} Similar to the previous discussion, we define a subset $\widetilde{\mathfrak{X}}_\mathcal{B}\subseteq\mathbb{R}^{n_1\times n_2\times n_3}$ as
\begin{equation}\label{PoissXB1B}
\widetilde{\mathfrak{X}}_\mathcal{B}:=\left\{\mathcal{X}
=(\mathcal{A}_0+\mathcal{A}_1)\diamond\mathcal{B}: \mathcal{B}\in\widetilde{\mathfrak{B}}_1\right\},	
\end{equation}
where
$$
\mathcal{A}_0:=(\mathcal{M}_1 \ \mathbf{0})\in\mathbb{R}^{n_1\times r\times n_3}  \ \text{with} \ \mathcal{M}_1\in\mathbb{R}^{n_1\times 1\times n_3},
$$
and
$$
\mathcal{A}_1:=
\begin{pmatrix}
\mathbf{0}_{r'1} & \mathcal{I}_{r'} & \mathbf{0} \\
\vdots & \vdots  & \vdots \\
\mathbf{0}_{r'1} &  \mathcal{I}_{r'} & \mathbf{0} \\
\mathbf{0}_{r'1} & \mathbf{0} & \mathbf{0} \end{pmatrix}\in\mathbb{R}^{n_1\times r\times n_3}.
$$
Here  $r'=\lceil\frac{s}{n_2n_3}\rceil-1$, $\mathbf{0}_{r'1}\in\mathbb{R}^{r'\times 1\times n_3}$ is a zero tensor,
$\mathcal{I}_{r'}\in\mathbb{R}^{r'\times r'\times n_3}$ is the identity tensor, where there are $\lfloor\frac{n_1}{r'}\rfloor$ blocks $\mathcal{I}_{r'}$ in $\mathcal{A}_1$,
$\mathcal{M}_1\in\mathbb{R}^{n_1\times 1 \times n_3}$ denotes a tensor  that the first frontal slice is all one and other frontal slices are zeros.
$
\widetilde{\mathfrak{B}}_1\subseteq\mathfrak{B}_1
$
is defined as
\begin{equation}\label{PoisXBsubB1}
\widetilde{\mathfrak{B}}_1:=\left\{\mathcal{B}=
\begin{pmatrix} \zeta\mathbb{I}_1 \\ \mathcal{B}_{r'}\\ \mathbf{0} \end{pmatrix},
\mathbb{I}_1\in\mathbb{R}^{1\times n_2\times n_3}, \mathcal{B}_{r'}\in \mathbb{R}^{r'\times n_2\times n_3},
(\mathcal{B}_{r'})_{ijk}\in\{0,b_0\},\|\mathcal{B}_{r'}\|_0\leq s-n_2n_3 \right\},	
\end{equation}
where  $\mathbb{I}_1$ represents a tensor with all entries being ones.
By the definition of tensor-tensor product and the structure of $\mathcal{A}_1$,
we get that
$\mathcal{A}_1\diamond\mathcal{B}=\mathcal{A}_1\diamond\mathcal{B}'$,
where
$$
\mathcal{B}'=
\begin{pmatrix} \mathbf{0}_1 \\ \mathcal{B}_{r'}\\ \mathbf{0} \end{pmatrix}
$$
with $\mathbf{0}_1\in\mathbb{R}^{1\times n_2\times n_3}$ being a zero tensor.
For any $\mathcal{X}\in\widetilde{\mathfrak{X}}_\mathcal{B}$, we have
\begin{equation}\label{EXXp}
\begin{split}
\mathcal{X}& =\mathcal{A}_0\diamond\mathcal{B}+\mathcal{A}_1\diamond\mathcal{B}\\
& = \textup{Fold}\left(\begin{pmatrix}
\mathbf{N}_{n_1r} &  \mathbf{0} & \cdots &  \mathbf{0} \\
 \mathbf{0}   & \mathbf{N}_{n_1r}  & \cdots &  \mathbf{0} \\
 \vdots            &  \vdots           &        \ddots & \vdots \\
\mathbf{0}  &\mathbf{0} &\cdots &  \mathbf{N}_{n_1r}  \end{pmatrix}\cdot
\begin{pmatrix}
\mathbf{B}^{(1)}  \\
\mathbf{B}^{(2)}   \\
 \vdots    \\
\mathbf{B}^{(n_3)}
\end{pmatrix} \right) + \mathcal{A}_1\diamond\mathcal{B}'\\
& =
\textup{Fold}
\begin{pmatrix} \zeta \mathbf{E}_{n_1n_2} \\  \zeta \mathbf{E}_{n_1n_2}
\\ \vdots \\ \zeta \mathbf{E}_{n_1n_2} \end{pmatrix} +\mathcal{A}_1\diamond\mathcal{B}' \\
& =\zeta \mathbb{I}_n+
\mathcal{A}_1\diamond\mathcal{B}',
\end{split}
\end{equation}
where $\mathbf{N}_{n_1r}= (\mathbf{E}_{n_11} \ \mathbf{0}_{n_1(r-1)})\in\mathbb{R}^{n_1\times r}$ with $\mathbf{E}_{n_11} \in\mathbb{R}^{n_1\times 1}$ being a  column vector (all $1$) and $\mathbf{0}_{n_1(r-1)}\in \mathbb{R}^{n_1\times (r-1)}$ being a zero matrix,
$$
\mathbf{B}^{(i)}=
\begin{pmatrix} \zeta \mathbf{E}_{1n_2}  \\ \mathbf{B}_{r'}^{(i)}\\ \mathbf{0} \end{pmatrix}
$$ with  $\mathbf{E}_{1n_2} \in\mathbb{R}^{1\times n_2}$ being a  row vector (all $1$) and $\mathbf{B}_{r'}^{(i)}$ being the $i$th frontal slice of $\mathcal{B}_{r'}$,
and $\mathbb{I}_n\in\mathbb{R}^{n_1\times n_2\times n_3}$ is a tensor with all entries being $1$.
Therefore,  $\mathcal{X}\in \widetilde{\mathfrak{U}}(r,b,s,\zeta)$, which implies that
$\widetilde{\mathfrak{X}}_\mathcal{B}\subseteq\widetilde{\mathfrak{U}}(r,b,s,\zeta)$.
Therefore, by applying the Varshamov-Gilbert
bound \cite[Lemma 2.9]{tsybakov2009} to the last term of (\ref{EXXp}),
for any $\mathcal{X}_1,\mathcal{X}_2\in\widetilde{\mathfrak{X}}_{\mathcal{B}}^0$,
there exists a subset
$\widetilde{\mathfrak{X}}_{\mathcal{B}}^0\subseteq\widetilde{\mathfrak{X}}_{\mathcal{B}}$
such that
$|\widetilde{\mathfrak{X}}_{\mathcal{B}}^0|\geq2^{\frac{s-n_2n_3}{8}}+1$
and
\[
\begin{split}
\|\mathcal{X}_1-\mathcal{X}_2\|_F^2&\geq \left(\frac{s-n_2n_3}{8}\right)\left\lfloor\frac{n_1}{r'}\right\rfloor b_0^2\\
&\geq \left(\frac{s-n_2n_3}{16}\right)\cdot \frac{n_1}{r'}\cdot
\min\left\{b^2, \ \beta_b^2\frac{\zeta}{\Delta_1}\frac{s-n_2n_3}{m}\right\}\\
& \geq \frac{n_1n_2n_3}{32}\Delta_1
\min\left\{b^2, \ \beta_b^2\frac{\zeta}{\Delta_1}\frac{s-n_2n_3}{m}\right\}\\
& = \frac{n_1n_2n_3}{32}\cdot
\min\left\{\Delta_1 b^2, \ \beta_b^2\zeta\frac{s-n_2n_3}{m}\right\},
\end{split}
\]
where the third inequality holds by the fact that $\frac{x}{\lceil x\rceil}\geq \min\{\frac{1}{2},x\}$ for any $x>0$.
The KL divergence of $p_{\mathcal{X}_\Omega}(\mathcal{Y}_{\Omega})$ from
$p_{(\mathcal{X}_0)_\Omega}(\mathcal{Y}_{\Omega})$ is
\[
\begin{split}
D(p_{\mathcal{X}_\Omega}(\mathcal{Y}_{\Omega})||p_{(\mathcal{X}_0)_\Omega}(\mathcal{Y}_{\Omega}))
&=\frac{m}{n_1n_2n_3}\sum_{i,j,k}D(p_{\mathcal{X}_{ijk}}(\mathcal{Y}_{ijk})||p_{(\mathcal{X}_0)_{ijk}}(\mathcal{Y}_{ijk})) \\
&\leq \frac{m}{n_1n_2n_3}\sum_{i,j,k}\frac{(\mathcal{X}_{ijk}-\zeta)^2}{\zeta}\\
&\leq m\frac{b_0^2}{\zeta}\Delta_1\leq \beta_b^2(s-n_2n_3)\leq  \frac{8\beta_b^2\log(|\mathfrak{X}_\mathcal{B}^0|-1)}{\log(2)},
\end{split}
\]
where the second inequality follows from
$\|\mathcal{A}_1\diamond\mathcal{B}'\|_0\leq \lfloor\frac{n_1}{r'}\rfloor(s-n_2n_3)\leq n_1n_2n_3\Delta_1$
and the last inequality follows from $|\widetilde{\mathfrak{X}}_{\mathcal{B}}^0|\geq2^{\frac{s-n_2n_3}{8}}+1$.
Therefore,
by choosing $0<\beta_b\leq \frac{\sqrt{\widetilde{\alpha}_2\log(2)}}{2\sqrt{2}}$ with $0<\widetilde{\alpha}_2<1/8$, we have
$$
\frac{1}{|\mathfrak{X}_\mathcal{B}^0|-1}\sum_{\mathcal{X}\in \mathfrak{X}_\mathcal{B}^0}
D(p_{\mathcal{X}_\Omega}(\mathcal{Y}_{\Omega})||p_{(\mathcal{X}_0)_\Omega}(\mathcal{Y}_{\Omega}))
\leq \widetilde{\alpha}_2\log(|\mathfrak{X}_\mathcal{B}^0|-1).
$$
By \cite[Theorem 2.5]{tsybakov2009}, we obtain that
\begin{equation}\label{PXKL2}
\begin{split}
& \inf_{\widetilde{\mathcal{X}}}\sup_{\mathcal{X}^*
\in\widetilde{\mathfrak{X}}_{\mathcal{B}}}\mathbb{P}\left(\frac{\|\widetilde{\mathcal{X}}-\mathcal{X}^*\|_F^2}{n_1n_2n_3} \geq
\frac{1}{64}\min\left\{\Delta_1 b^2, \ \beta_b^2\zeta\frac{s-n_2n_3}{m}\right\}\right)\\
\geq &  \inf_{\widetilde{\mathcal{X}}}\sup_{\mathcal{X}^*
\in\widetilde{\mathfrak{X}}_{\mathcal{B}}^0}\mathbb{P}\left(\frac{\|\widetilde{\mathcal{X}}-\mathcal{X}^*\|_F^2}{n_1n_2n_3} \geq\frac{1}{64}\min\left\{\Delta_1 b^2, \ \beta_b^2\zeta\frac{s-n_2n_3}{m}\right\}\right)\geq \widetilde{\theta}_2,
\end{split}
\end{equation}
where
$$
\widetilde{\theta}_2=\frac{\sqrt{|\widetilde{\mathfrak{X}}_{\mathcal{B}}^0|-1}}
{1+\sqrt{|\widetilde{\mathfrak{X}}_{\mathcal{B}}^0|-1}}\left(1-2\widetilde{\alpha}_2
-\sqrt{\frac{2\widetilde{\alpha}_2}{\log(|\widetilde{\mathfrak{X}}_{\mathcal{B}}^0|-1)}}\right)\in(0,1).
$$
By combining (\ref{PXKL}) and (\ref{PXKL2}), we deduce
$$
\inf_{\widetilde{\mathcal{X}}}\sup_{\mathcal{X}^*
\in\mathfrak{U}(r,b,k)}\mathbb{P}\left(\frac{\|\widetilde{\mathcal{X}}-\mathcal{X}^*\|_F^2}{n_1n_2n_3} \geq\frac{3}{128} \min\left\{\widetilde{\Delta} b^2,\widetilde{\beta}_c^2\zeta\left(\frac{s-n_2n_3+rn_1n_3}{m}\right)\right\}\right)\geq \widetilde{\theta}_c,
$$
where $\widetilde{\Delta}:=\min\{(1-\varsigma)^2, \Delta_1\}$,
$\widetilde{\beta}_c:=\min\{\beta_a,\beta_b\}$,
 and $\widetilde{\theta}_c=\min\{ \widetilde{\theta}_1, \widetilde{\theta}_2\}$.
 By Markov's inequality, the desired conclusion is obtained easily.

%
%
%
%

\ifCLASSOPTIONcaptionsoff
  \newpage
\fi



%
%
%

\bibliographystyle{IEEEtran}
\bibliography{RefPNTFC}

%

%
%
%




\end{document}